\documentclass[acmsmall,screen]{acmart}
\AtBeginDocument{%
  \providecommand\BibTeX{{%
    \normalfont B\kern-0.5em{\scshape i\kern-0.25em b}\kern-0.8em\TeX}}}

\setcopyright{acmcopyright} 
\acmJournal{CSUR} 
\acmYear{2024}
\acmDOI{10.1145/3707446}
\acmMonth{11} 
\acmPrice{}

\usepackage{colortbl}
\usepackage{graphicx}
\usepackage{array,multirow}
\usepackage{threeparttable}
\usepackage{enumitem}
\usepackage{makecell}
\usepackage{pifont}
\usepackage{graphics} 
\usepackage{epsfig} 
\usepackage{amsmath} 

\usepackage{amssymb} 
\usepackage{caption}
\usepackage{multirow} 
\usepackage{color}
\usepackage{makecell}
\usepackage{subfigure}
\usepackage{lineno}
\usepackage{fancyhdr}

\newcommand{\etal}{\textit{et al.}}

\begin{document}
\fancyfoot[RO,LE]{\footnotesize ACM Computing Surveys} 
	

\title{LiDAR-Based Place Recognition For Autonomous Driving: A Survey}

\author{Yongjun Zhang}
\email{zhangyj@whu.edu.cn}
  \affiliation{%
  \institution{Wuhan University}
  \city{Wuhan 430072}
  \country{China}}
\author{Pengcheng Shi}
\email{shipc_2021@whu.edu.cn}
  \affiliation{%
  \institution{Wuhan University}
  \city{Wuhan 430072}
  \country{China}}
\author{Jiayuan Li}
\email{ljy_whu_2012@whu.edu.cn}
  \affiliation{%
  \institution{Wuhan University}
  \city{Wuhan 430072}
  \country{China}}
\thanks{Corresponding authors: Pengcheng Shi, Jiayuan Li}

\begin{abstract}
LiDAR has gained popularity in autonomous driving due to advantages like long measurement distance, rich 3D information, and stability in harsh environments. Place Recognition (PR) enables vehicles to identify previously visited locations despite variations in appearance, weather, and viewpoints, even determining their global location within prior maps. This capability is crucial for accurate localization in autonomous driving. Consequently, LiDAR-based Place Recognition (LPR) has emerged as a research hotspot in robotics. However, existing reviews predominantly concentrate on Visual Place Recognition (VPR), leaving a gap in systematic reviews on LPR. This paper bridges this gap by providing a comprehensive review of LPR methods, thus facilitating and encouraging further research. We commence by exploring the relationship between PR and autonomous driving components. Then, we delve into the problem formulation of LPR, challenges, and relations to previous surveys. Subsequently, we conduct an in-depth review of related research, which offers detailed classifications, strengths and weaknesses, and architectures. Finally, we summarize existing datasets and evaluation metrics and envision promising future directions. This paper can serve as a valuable tutorial for newcomers entering the field of place recognition. We plan to maintain an up-to-date project on https://github.com/ShiPC-AI/LPR-Survey.
\end{abstract}



\maketitle
\section{Introduction}\label{sec:introduction}
\subsection{Background}\label{sec:background}
In recent years, autonomous driving has rapidly advanced with applications in logistics, public transportation, food delivery, warehousing, and medical rescue. The autonomous driving system integrates complex modules such as sensors, perception, localization, planning, control, and communication. These modules employ techniques like computer vision, sensor technology, and data processing to enable vehicle autonomy. Location information is a prerequisite for the entire system, helping vehicles understand their environment for better navigation. Notably, Place Recognition (PR)~\cite{2015-TPAMI-VPRW,2019-ICCV-LpdNet,2022-AAAI-SVTNet,2021-TRO-ScanContext++,2024-TIV-ANHE} is a vital technique for obtaining and expressing this information.
\begin{figure*}[!t]
	\centering
	\includegraphics[width = 12cm]{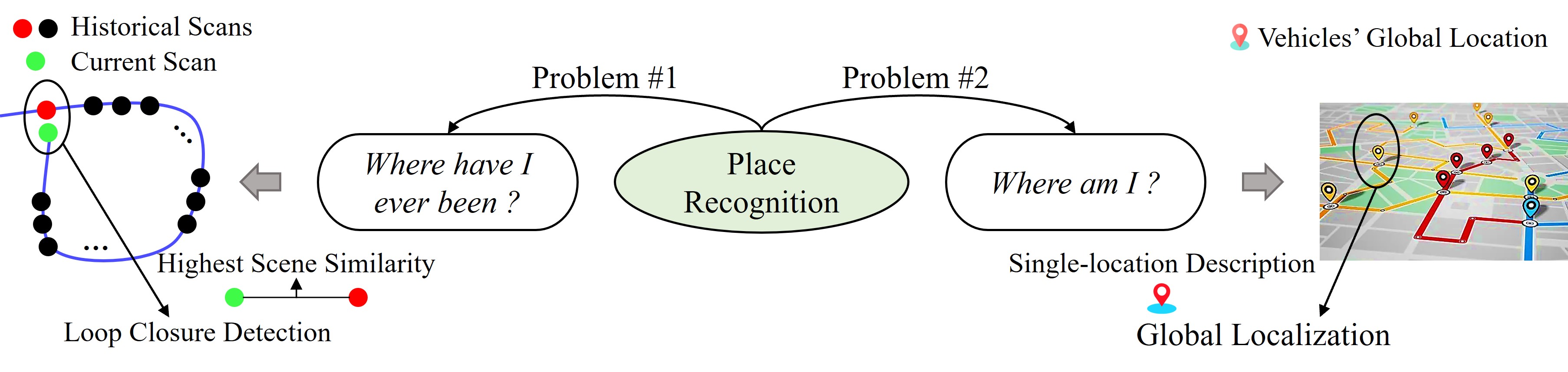}
	\caption{Two key problems addressed by PR. On the left, the blue line denotes the vehicle trajectory, with solid circles indicating collected scans by sensors over time. The green circle marks the current scan, while the others represent past scans. The green and red circles are geographically close, and their scans exhibit the highest scene similarity, forming a closed loop. On the right, the black-marked area shows a vehicle's global location, only providing a single-location description within maps.}\label{fig:pr-problem}
\end{figure*}

As shown in Figure \ref{fig:pr-problem}, we summarize PR's functions into two aspects. (1) It addresses the problem of ''\textit{\textbf{where have I ever been}}'', also known as Loop Closure Detection (LCD)~\cite{2022-TITS-Survey-TRPI,2020-ICRA-ISC,2011-IROS-NARF}. When the robot explores unknown environments, PR compares newly captured sensor data with historical data to identify revisited locations. The loop closure data is then passed to the backend system to establish constraints and mitigate pose drifts. In this context, PR and localization are interdependent, with PR enhancing localization accuracy through loop closure detection and error correction. (2) It tackles the issue of "\textit{\textbf{where am I}}", also known as global localization~\cite{2020-CSUR-AVNF,shi2024indoor}. When the vehicle travels within a predefined map, PR correlates newly captured sensor data with maps to pinpoint the vehicle's locations. Subsequently, the planning module utilizes this location to optimize path prediction, while the control module generates corresponding vehicle commands. In this context, PR is a specialized localization method that directly provides the vehicle's global pose.

To date, place recognition remains a challenging and ongoing research problem. Traditional Global Positioning System (GPS)~\cite{2009-TITS-RHOS} provides absolute positions but is easily blocked in underground parking lots and canyons, limiting their effectiveness. Inertia Measurement Unit (IMU)~\cite{xia2021autonomous} offers high-frequency attitude data but suffers from accuracy degradation over time due to cumulative errors. Wheel odometers~\cite{2021-TRO-PEFG} provide stable displacement information, but slippery roads or sudden acceleration can distort data. In contrast, measurement sensors like cameras and LiDAR provide high-resolution environmental information with frequent updates. Cameras~\cite{moller2014illumination} excel in capturing visual details like color, shape, and texture but are susceptible to illumination changes and a narrow Field Of View (FOV). In contrast, LiDAR~\cite{2024-TPR-3LSA} generates rich 3D data and operates reliably in harsh conditions (e.g., night, rain, fog), signal-limited environments, and complex terrain. These attributes greatly enhance LiDAR's suitability for autonomous driving applications, leading to growing interest in LiDAR-based Place Recognition (LPR).
\subsection{Relation to Autonomous Driving}
\subsubsection{Overview of Autonomous Driving Components}
Autonomous driving~\cite{badue2021self} enables vehicles to execute tasks independently using sensors, computer systems, and artificial intelligence, managing acceleration, braking, and steering while ensuring safety and efficiency. The American Society of Autonomous Driving Engineers (SAE) classifies autonomous driving into six levels~\cite{sae2018taxonomy}, ranging from full human control (L0) to complete automation (L5). The key to vehicle automation and intelligence is ensuring the core system components effectively interact and collaborate. The industry and academia generally adopt a modular approach, viewing autonomous driving systems as extensions of mobile robot architectures~\cite{2020-TNNLS-ASOE}, including perception, localization, planning, control, and Human-Machine Interface (HMI).
	
\textbf{Perception}. The perception module ~\cite{zablocki2022explainability} acquires real-time environmental data through multiple sensors. It performs target detection~\cite{2016-TITS-FDFV}, classification~\cite{2016-CVPR-VolumetricCNN}, and tracking~\cite{2020-IROS-LiTAMIN} to provide high-precision environmental information for localization, planning, and control. It identifies traffic signs, lanes, and lights to evaluate road conditions. Additionally, it detects and tracks obstacles to determine their proximity to maintain a safe driving distance.

\textbf{Localization}. The localization module ~\cite{2017-CSUR-GFSI} determines the vehicle's position to support navigation, decision, and control. It typically relies on multiple sensors, including LiDAR, cameras, GPS, and IMU, to manage complex scenarios. GPS provides location data in open areas, while High-Definition (HD) maps ~\cite{2022-ISPRSJ-PGHCI} offer detailed road information in GPS-denied environments, helping anticipate hazards like sharp turns and steep slopes. IMU~\cite{xia2021autonomous} measures acceleration and angular velocity to help infer the vehicle's motion in high-speed scenarios.

\textbf{Planning}. The planning module~\cite{claussmann2019review} connects perception, localization, and control, which employs path optimization and decision-making technologies to develop smooth and safe driving routes in complex environments. It establishes a global path from the start to the destination based on criteria such as shortest distance or minimum energy consumption. Simultaneously, it dynamically adjusts the local path using real-time perception data and vehicle location to avoid obstacles and address emergencies.

\textbf{Control}. The control module~\cite{2023-ESWA-ADSA}, interfacing directly with the vehicle, employs perception data, vehicle location, and planning instructions to generate operational commands. It controls acceleration, braking, and steering to guide vehicles to follow the predefined and dynamically adjusted paths, ensuring smooth and safe driving. Additionally, it receives perception data and location information to manage honking and lighting to alert pedestrians and other vehicles to avoid accidents.

\textbf{Human-Machine Interface}. The HMI module ~\cite{gruyer2017perception} facilitates communication between users and vehicles, enhancing the transparency of autonomous driving systems and user experience. It enables interactions through voice, touch, or gesture, allowing users to convey instructions and understand system responses. By providing intuitive interfaces and feedback, HMI helps users grasp system status, warnings, and decisions, offering clear guidance at critical moments to ensure safety and reliability in diverse scenarios.
\begin{figure*}[!t]
	\centering
	\includegraphics[width = 13cm]{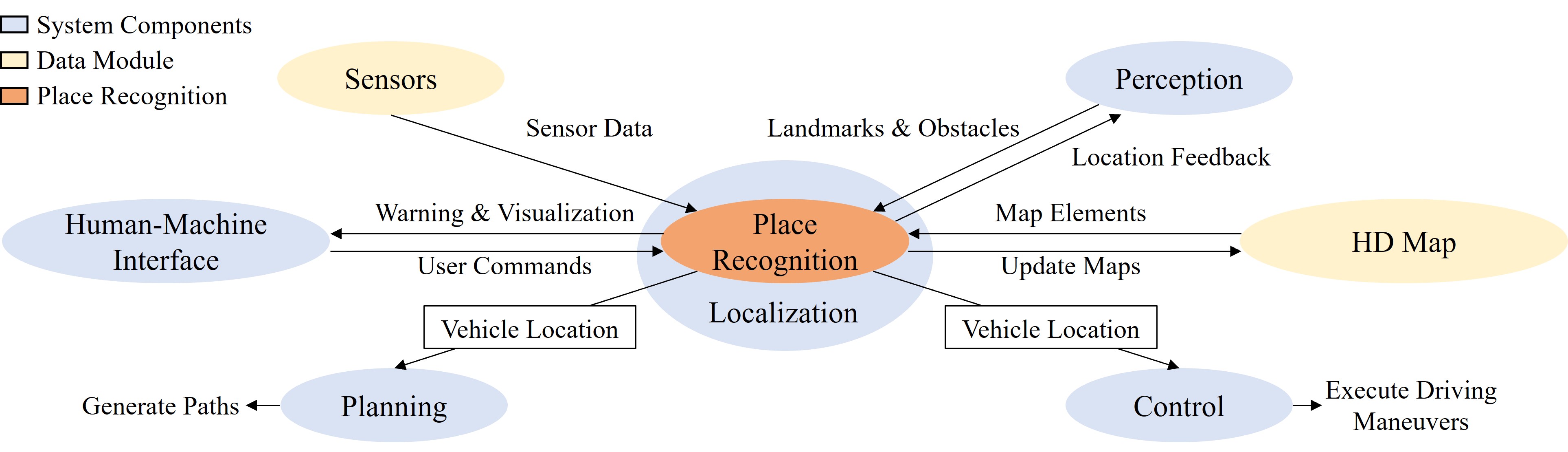}
	\caption{The relationship between PR and key components of autonomous driving. HD maps, being external information, are grouped with sensors under data modules.}\label{fig:relation-autonomous-driving}
\end{figure*}
\subsubsection{Relationship Between PR and Each Component}
As illustrated in Figure \ref{fig:relation-autonomous-driving}, we summarize the relationship between PR and autonomous driving components:

\textbf{High-Precision Localization}. In autonomous driving systems with HD maps, PR is typically viewed as a component of the localization module. It integrates sensor data~\cite{2016-TIE-RVRA}, perceived landmarks~\cite{2021-ECMR-ORIP}, and HD map elements~\cite{2022-arXiv-Survey-ASOV} to accurately pinpoint the vehicle's location, even in GPS-denied areas. HD maps, in turn, use the vehicle's location and sensor data to monitor environmental changes and update the map if significant alterations are detected.

\textbf{Environment Perception Support}. PR and perception modules exchange information and feedback with each other, facilitating data fusion and error correction to enhance perception and localization accuracy~\cite{chen2015deepdriving}. The perception identifies landmarks like lanes and traffic signs, supplying key features for PR to use in localization. PR offers location feedback to the perception, comparing real-time data with maps to correct perception errors.

\textbf{Path Planning Foundation}. The precise location data from PR aids the planning module in selecting the optimal path based on current road conditions and adjusting it for real-time traffic changes~\cite{bachute2021autonomous}. It also supports precise obstacle avoidance and lane-keeping in complex scenarios, ensuring safe vehicle operation.

\textbf{Accurate Control Assistance}. PR provides the control module with crucial location data, enabling precise control commands and real-time adjustments~\cite{kalaria2023delay}. This data helps the module determine the vehicle's position and direction, allowing accurate acceleration, braking, and steering to follow the planned path. If deviating from the trajectory, the control module swiftly corrects the course to keep vehicles on the correct route.

\textbf{Enhanced Human-Machine Interactivity}. PR aids the HMI in showing the vehicle's precise location, route, and destination, which enhances user understanding of the vehicle's status~\cite{omeiza2021explanations}. In abnormal situations, HMI employs PR's feedback to issue warnings or provide suggestions, which helps users deal with current driving conditions. Users can enter destinations, preferences, and modes through HMI, and PR updates navigation and optimizes strategies based on real-time data, improving the user experience.

Therefore, PR is crucial in autonomous driving, supporting localization, perception, planning, control, and HMI modules by providing accurate vehicle location data. It boosts localization accuracy, enables the system to adapt to complex environments, ensures efficient and safe vehicle operation, and improves user experience. Effective collaboration among these modules creates a comprehensive solution that allows vehicles to navigate safely in diverse and dynamic scenarios.
\subsection{Contributions}\label{sec:contributions-and-paper-organization}
\begin{figure*}[t]
	\centering
	\includegraphics[width = 12.5cm]{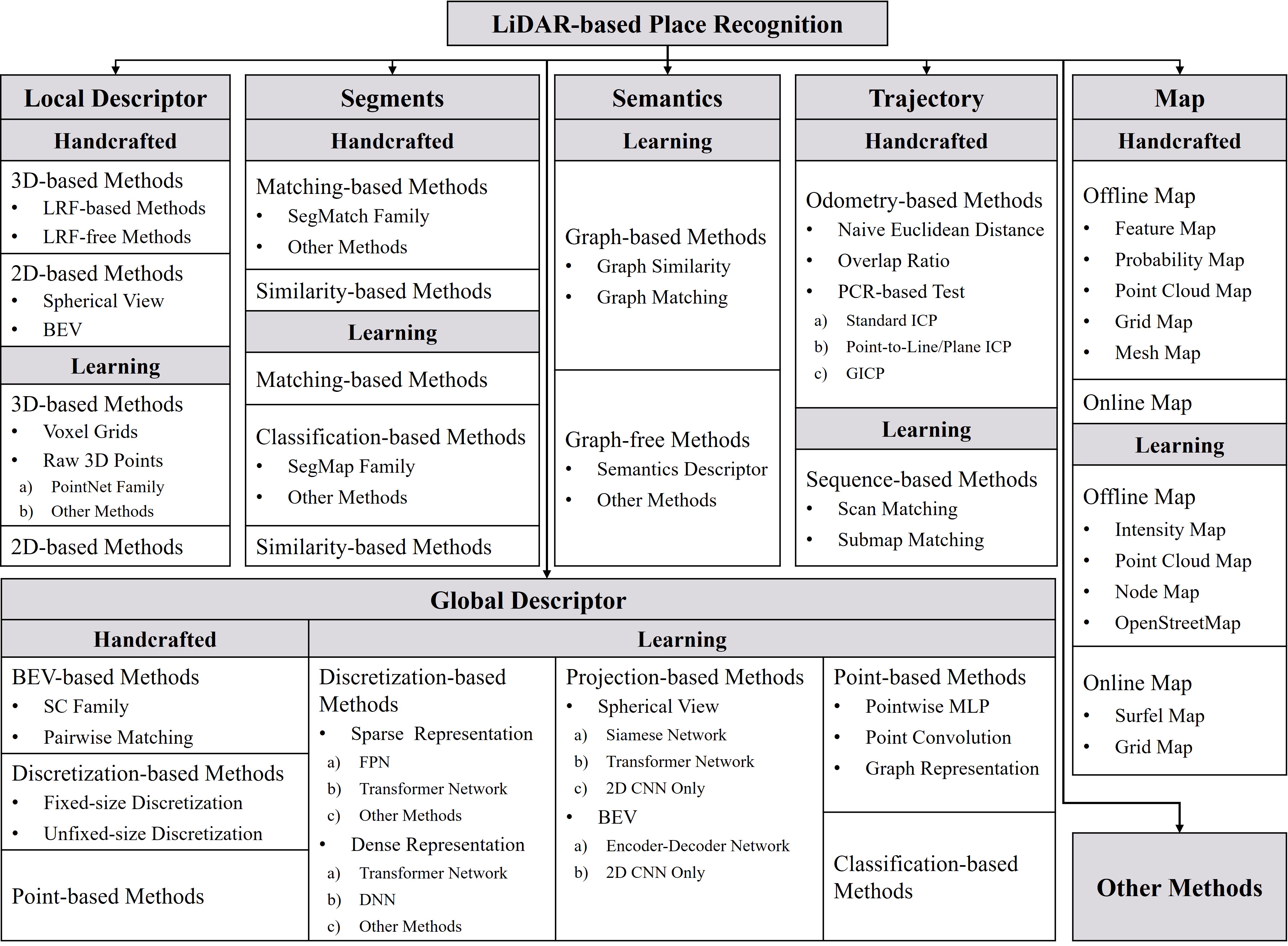}
	\caption{Proposed taxonomy of LPR methods.}\label{fig:overview-methods}
\end{figure*}
In this paper, we present a comprehensive review of LPR research, accompanied by a detailed methodological taxonomy depicted in Figure \ref{fig:overview-methods}. We categorize methods into handcrafted and learning-based types, further subdividing them, and present detailed introductions to pioneering works. This well-organized layout enhances reading efficiency and facilitates a better understanding of the relevant technologies in place recognition. Our main contributions are as follows:
\begin{itemize}
	\item \textbf{Filling Survey Gap}. Existing reviews focus on VPR or discuss general place recognition issues with little emphasis on LiDAR. Our dedicated survey on LPR bridges this gap, helping researchers understand and disseminate State-Of-The-Art (SOTA) LPR methods.
	\item \textbf{Detailed Definition and Method Classification}. We offer a comprehensive definition of place recognition, including implicit loop closure detection and explicit global localization. The proposed taxonomy of LPR methods includes seven parts, each split into handcrafted and learning-based branches, with further subcategories based on specific rules. We also outline their advantages and disadvantages. The efforts may help researchers understand the applicability of methods and inspire the development of place recognition techniques for challenging scenarios like docks, parks, and woodlands.
	\item \textbf{Future Direction}. We propose promising directions for advancing LPR research: innovative solutions like cloud computing, quantum technology, bionic localization, and applications in space exploration, polar research, and underwater robotics. These areas remain unexplored in previous surveys. We also highlight multimodal information sources such as WiFi, voice, and Radio Frequency Identification (RFID), along with advanced sensors like solid-state lidar, event cameras, and millimeter-wave radar. Additionally, we recommend evaluating methods based on scale, efficiency, long-term performance, and developing standard datasets.
	\item \textbf{Open-source Project}. We maintain an up-to-date project on our website, allowing the robotics community to stay current with SOTA LPR technologies. This resource helps newcomers quickly grasp essential information such as datasets, evaluation metrics, and mainstream methods.
\end{itemize}
\section{Definition and Challenges}\label{sec:definition-and-challenges}
\subsection{Problem Formulation}\label{sec:problem-formulation}
\begin{figure*}[t]
	\centering
	\includegraphics[width = 12cm]{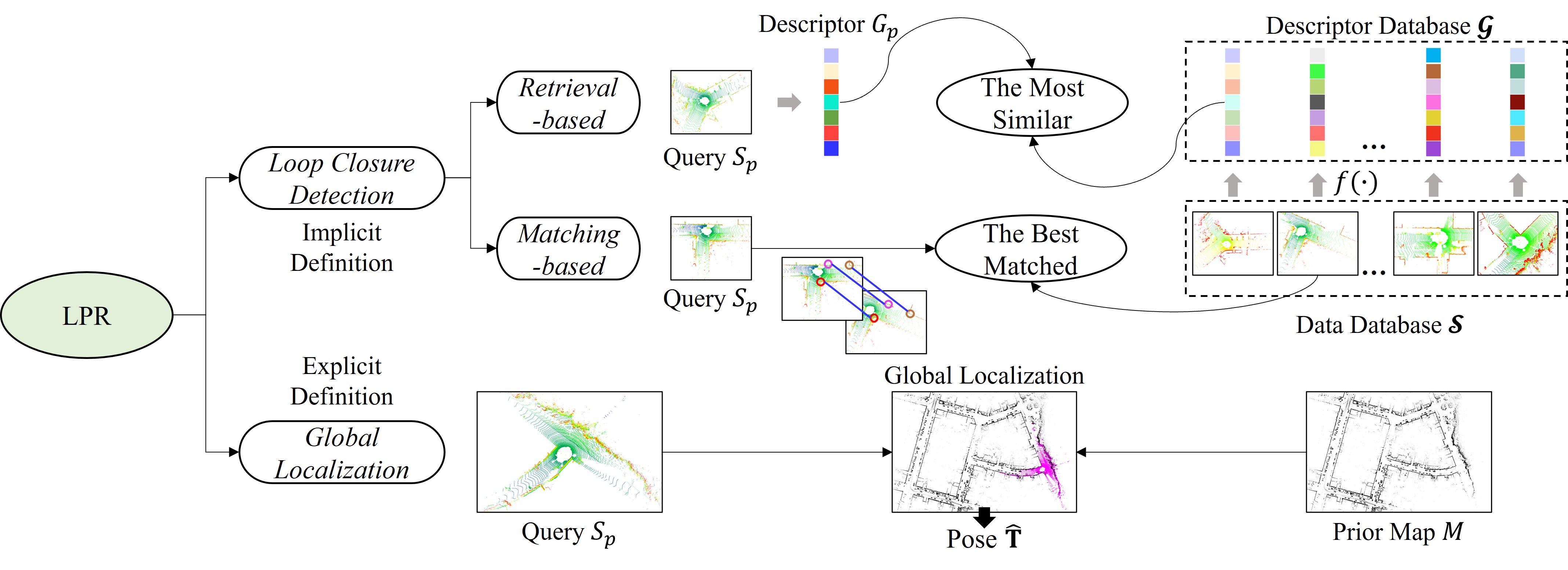}
	\caption{Definition of LiDAR-based place recognition. We classify the definition into implicit loop closure detection and explicit global localization. Since global localization provides global vehicle poses, we term it explicit place recognition. In contrast, loop closure detection identifies revisited poses through frame-to-frame comparison without a global pose, making it implicit.}\label{fig:pr-definition}
\end{figure*}

As summarized in ~\cite{2022-arXiv-Survey-GPRS}, PR has two prevalent definitions: overlap-based and distance-based. The overlap-based definition~\cite{2021-PR-Survey-VPRA} emphasizes visual similarity, identifying two images depicting the same place if they show significant visual overlap. However, the distance-based definition~\cite{2022-arXiv-Survey-ASOV} relies on geographical proximity, defining places as identical if the distance between two vehicles is below a user-defined threshold. As our survey focuses on LiDAR sensors, we follow the distance-based definition. Differently, we present two definitions for LPR: implicit loop closure detection and explicit global localization. As shown in Figure \ref{fig:pr-definition}, loop closure detection solves the revisit problem by determining if two locations are close based on similarity or matching ratio. Due to the absence of direct global poses, it is called implicit place recognition. In contrast, global localization provides the vehicle's poses within a pre-existing map, termed explicit place recognition.
\subsubsection{Loop Closure Detection}
We follow the definition in ~\cite{2022-arXiv-Survey-GPRS}. Let $S_p$ denote the current sensor data, $L_p$ the current robot location, $\mathcal{L} = \{L_1, L_2, \ldots, L_k\}$ previously visited locations, and $\mathcal{S} = \{S_1, S_2, \ldots, S_k\}$ previously collected data, with subscripts indicating the data index. Given another sensor data $S_q \in \mathcal{S}$ collected at location $L_q \in \mathcal{L}$, a loop closure occurs when $L_p$ and $L_q$ are geographically close. Two geographically close locations typically have similar environmental layouts, which result in their sensor data with high matching ratios:
\begin{equation}
	dist(L_p, L_q) < \lambda_l, \ \ score(S_p, S_q) > \lambda_s 
	\label{eq:lcd}
\end{equation}
where $dist(L_p, L_q)$ denotes the geographical distance between $L_p$ and $L_q$, $score(S_p, S_q)$ means the matching ratio for $S_p$ and $S_q$, $\lambda_l$ and $\lambda_s$ are two user-defined thresholds based on a specific application. Identifying the closest location is then transformed into searching the data with the highest similarity score to the current data:
\begin{equation}
	\widehat{L}_q = \underset{L_q \in \mathcal{L}}{\arg\min}\ dist(L_p, L_q) \Rightarrow \widehat{S}_q = \underset{S_q \in \mathcal{S}}{\arg\max}\ score(S_p, S_q),
	\label{eq:definition-pr-lcd}
\end{equation}
where $\widehat{L}_q$ is the closest location, and $\widehat{S}_q$ represents the most similar or best-matched sensor data.

Some researchers~\cite{2018-CVPR-Pointnetvlad,2019-CVPR-PCAN,2020-ICMR-DAGC,2018-IROS-ScanContext,2021-TRO-ScanContext++} treat the above searching for best-matched data as a retrieval task. These methods typically encode sensor data into a global descriptor $G$ and aggregate historical descriptors into a database $\mathcal{G}$. During retrieval, the current descriptor $G_p$ is matched against the database $\mathcal{G}$ to find the most similar descriptor $\widehat{G}_q$:
\begin{equation}
	\widehat{L}_q = \underset{L_q \in \mathcal{L}}{\arg\min}\ dist(L_p, L_q) \Rightarrow \widehat{G}_q = \underset{G_q \in \mathcal{G}}{\arg\min}\ \delta(G_p, G_q), G_p = f(S_p), G_q = f(S_q),
	\label{eq:definition-pr-lcd-retrieval}
\end{equation}
where $\delta(\cdot)$ denotes the descriptor distance, typically the $L_2$ norm, $f(\cdot)$ represents a descriptor encoding process. Since encoding methods vary widely without a unified rule, we do not provide a specific definition. Unlike image retrieval~\cite{2019-TPAMI-GeM} in computer vision, this operates directly on point clouds and uses the geographical distance between vehicles as the metric.
\subsubsection{Global Localization}
Since our survey focuses on LiDAR sensors, our definition of global localization excludes GPS-based methods~\cite{2012-TITS-GPS-L}. Let $S_p$ represent the current sensor data and $M$ represent a prior map. Global localization~\cite{2022-RAL-ESSO} establishes associations between $S_p$ and $M$ to pinpoint the vehicle's global pose:
\begin{equation}
	\widehat{\mathbf{T}} = \underset{\mathbf{T}}{\arg\min}\ loss(\mathbf{T}, S_p, M),
	\label{eq:definition-gl}
\end{equation}
where $\mathbf{T}$ is a transformation with six or three Degree-Of-Freedoms (DoF), $loss$ denotes the loss function between the current sensor data $S_p$ and the map data $M$ calculated based on the pose $\mathbf{T}$, generally the point-to-point or point-to-surface distance.
\subsection{Challenges of LPR}\label{sec:challenges-of-lpr}
Numerous recent LPR methods have explored techniques like Bird-Eye-View (BEV)~\cite{2018-IROS-ScanContext,2020-IROS-LiDAR-Iris}, histograms~\cite{2015-IROS-FastHistogram,2022-TIV-HOPN}, image representations~\cite{2019-IROS-OREOS,2019-RAL-SCI}, and graph theory~\cite{2020-IROS-SGPR, 2020-IROS-GosMatch} to enhance performance. They achieve rotational invariance through height similarity~\cite{2018-IROS-ScanContext,2020-ICRA-ISC} and frequency domain analysis~\cite{2020-IROS-LiDAR-Iris,2021-RAL-BVMatch}. Additional approaches employ pose proximity~\cite{2020-IROS-LIO-SAM}, sequence matching~\cite{2022-TIE-SequenceMatch}, and Point Cloud Registration (PCR)~\cite{2023-TPAMI-QGORE} techniques to improve recognition accuracy. However, these methods struggle in dynamic and highly occluded environments. Traditional methods rely on low-level features (coordinates~\cite{2013-ICRA-3DGestalt}, normals~\cite{2011-SSRR-Z-Projection}, intensities~\cite{2019-RAL-ISHOT}, and range~\cite{2020-ICRA-OneShot}), while learning-based approaches gradually show promising results using neural networks~\cite{ 2022-TIP-EPC-Net}, attention mechanisms~\cite{ 2020-ECCV-DH3d}, and semantics~\cite{2022-AR-OverlapNet}. Furthermore, diverse map representations, such as point clouds~\cite{2022-ISPRSJ-CSSC}, semantics~\cite{2019-IROS-Suma++}, and mesh~\cite{2021-ICRA-RILL}, have been successfully applied in map localization. Despite the impressive results claimed by these methods, several challenges persist that require further attention: 

\textbf{Motion Distortions}. Vehicle motion inevitably distorts the point cloud, severely affecting feature matching and registration between scans~\cite{2024-RAL-VICET}. Several methods~\cite{2021-IROS-F-LOAM, 2021-ICRA-MULLS} employ a constant velocity motion model based on the previous pose to correct this. Although effective in most scenarios, this approach is inadequate for sudden direction changes~\cite{2022-ICRA-CT-ICP} or rapid acceleration~\cite{2020-TRO-IN2LAAMA}.

\textbf{Viewpoint Differences}. Lane-level horizontal deviations may exist when a robot revisits a historical place from different directions. While a few methods~\cite{2018-IROS-ScanContext, 2019-RAL-SCI, 2020-ICRA-ISC} address rotation invariance, they overlook the impact of translation on place recognition.

\textbf{Weather Conditions}. Laser signals exhibit varying behavior under different weather conditions~\cite{2020-Camera-LiDAR-Comp}. They attenuate less and travel farther on sunny days but decay significantly in rainy and foggy weather.

\textbf{Perceptual Aliasing}. Distinct places in confined corridors~\cite{2021-AGCS-LIFM,2023-JAG-100FPS} may exhibit similar point cloud data, which introduces ambiguous interpretations.

\textbf{Appearance Changes}. Long-term navigation applications~\cite{2019-RAL-SCI,2021-ECMR-ORIP} often involve significant environmental changes, which leads to potential failures. 

\textbf{Sensor Characteristics}. Mechanical LiDAR~\cite{2018-IROS-Lego-LOAM} produces point clouds in a format of multiple scan lines, resulting in vertical sparsity. Solid-state LiDAR~\cite{2019-arXiv-Fast-Complete} provides limited horizontal Field Of View (FOV) and requires specific considerations.

\subsection{Relation to Previous Surveys}\label{sec:relation-to-previous-surveys}
In recent years, there has been a proliferation of reviews on visual technologies, addressing various topics like place recognition~\cite{2017-Robotics-VPRF, 2021-PR-Survey-VPRA, 2016-TRO-Survey-VPRA, 2022-TITS-Survey-TRPI},  localization~\cite{2021-arXiv-survey-VAOG,2022-arXiv-Survey-ASOV}, tracking~\cite{2022-CVIU-survey-VOTA,2021-TITS-survey-FLDV}, and SLAM~\cite{2022-ESA-survey-vSLAM-ASOS,2022-CSR-survey-vSLAM-VSFU,2023-arXiv-survey-vSLAM-ESLA}. Despite their valuable contributions to the progress of research in robotics and autonomous driving, these reviews have unfortunately overlooked the technology of LiDAR. Cadena \etal~\cite{2016-TRO-Survey-PPAF} extensively reviewed the current state of SLAM and delved into potential future directions. Yin \etal~\cite{2022-arXiv-Survey-GPRS} provided a comprehensive place recognition survey encompassing cameras, LiDAR, radar, and joint sensors. However, the section dedicated to discussing LiDAR was relatively limited. Yin \etal~\cite{2023-arXiv-survey-lgl-ASOG} offered an informative overview of the recent progress in LiDAR-based global localization, while it merely represented a specialized branch of place recognition. 

Comparatively, our survey distinguishes itself through the following features: (1) It is the first survey dedicated solely to LPR research, filling a gap in the field and advancing the dissemination of SOTA LPR methods. (2) It offers a more comprehensive problem formulation, method classification, and summary, enhancing understanding of methodological advantages, disadvantages, and applicability. The efforts facilitates the development of LPR techniques, especially in challenging scenarios. (3) We envision promising future directions to advance LPR research, including innovations such as cloud computing, quantum technology, and bionic localization, as well as applications in space exploration, polar research, and underwater robotics. (4) We maintain an up-to-date project to keep the robotics community updated on SOTA LPR technologies and aid newcomers in quickly grasping essential information like datasets, evaluation metrics, and methodologies.
\begin{table*}[tbp]
	\tiny
	\centering
	\caption{A summary of local descriptor-based methods.}
	\begin{tabular}{|m{1cm}<{\centering}|c|r|c|c|c|c|c|c|}
		\hline
		\rowcolor{gray!20} 
		\multicolumn{9}{|c|}{\textbf{Handcrafted Methods}}\\\hline
		\multicolumn{3}{|c|}{\textbf{Methods}}&\textbf{Year} &\textbf{Feature} &\textbf{Size} &\multicolumn{2}{c|}{\textbf{Similarity Metric}} & \textbf{Code}\\\hline
		\multirow{10}{*}{\makecell[c]{3D-based\\ Methods}}&\multirow{5}{*}{LRF-free} &Spin Image~\cite{1999-TPAMI-SpinImage}&1999&Density&153&\multicolumn{2}{c|}{L2 Distance}&\checkmark\\\cline{3-9}
		&&3DGestalt~\cite{2013-ICRA-3DGestalt}&2013&Height &32$\times$10 &\multicolumn{2}{c|}{Voting}&\\\cline{3-9}
		&&NBLD~\cite{2016-ICRA-NBLD} &2016 &Density &16$\times$4$\times$8 &\multicolumn{2}{c|}{Voting}&\\\cline{3-9}
		&&GLAROT-3D~\cite{2017-IROS-GLAROT-3D}&2017&Orientation+Range &1880 &\multicolumn{2}{c|}{Rotated L1 Norm}&\\\cline{3-9}
		&&HoPPF~\cite{2020-PR-HOPPF}&2020&Angle+Distance&600&\multicolumn{2}{c|}{}&\\\cline{2-9}
		&\multirow{5}{*}{LRF-based}&USC~\cite{2010-ACM-USC}&2010&Density&1960&\multicolumn{2}{c|}{Euclidean Distance}&\checkmark\\\cline{3-9}
		&&RoPS~\cite{2013-IJCV-ROPS}&2013&Density&135&\multicolumn{2}{c|}{L2 Distance}&\checkmark\\\cline{3-9}
		&&TOLDI~\cite{2017-PR-TOLDI}&2017&Depth&3$\times$20$\times$20&\multicolumn{2}{c|}{}&\\\cline{3-9}
		&&ISHOT~\cite{2019-RAL-ISHOT}&2019&Angle+Intensity&1344&\multicolumn{2}{c|}{Voting}&\checkmark\\\cline{3-9}
		&&Sun \etal~\cite{2020-IS-AEAC}&2020&Height&20$\times$20&\multicolumn{2}{c|}{Euclidean Distance}&\\\hline
		\multirow{7}{*}{\makecell[c]{2D-based\\ Methods}}&\multirow{5}{*}{Spherical View}&Steder \etal~\cite{2010-ICRA-RPRF}&2010&Range+Curvature&&\multicolumn{2}{c|}{Euclidean Distance}&\\\cline{3-9}
		&&Steder \etal~\cite{2011-IROS-NARF}&2011&Range+Curvature &36 &\multicolumn{2}{c|}{Manhattan Distance}&\\\cline{3-9}
		&&Zhuang \etal~\cite{2013-TIM-3SMA}&2013 &Space&&\multicolumn{2}{c|}{Matching Score}&\\\cline{3-9} 
		&&Cao \etal~\cite{2018-IEEESensors-RPCA}&2018&Position &600$\times$391 &\multicolumn{2}{c|}{L1-Norm}&\\\cline{3-9}
		&&Shan \etal~\cite{2021-ICRA-AIL} &2021 &Intensity &1024$\times$128 &\multicolumn{2}{c|}{L1 Distance+Hamming Distance}&\checkmark \\\cline{2-9}
		&\multirow{2}{*}{BEV}&BVMatch~\cite{2021-RAL-BVMatch}&2021 &Density&6$\times$6$\times$6&\multicolumn{2}{c|}{2D Rigid Pose} &\checkmark\\\cline{3-9}
		&&HOPN~\cite{2022-TIV-HOPN}&2022&Normal+Density&6$\times$6$\times$6 &\multicolumn{2}{c|}{2D Rigid Pose}&\checkmark\\\hline
		\rowcolor{gray!20}\multicolumn{9}{|c|}{\textbf{Learning-based Methods}}\\\hline
		\multicolumn{3}{|c|}{\textbf{Methods}}&\textbf{Year}&\textbf{Backbone}&\textbf{Size}&\textbf{Loss}&\textbf{EtE} &\textbf{Code}\\\hline
		\multirow{15}{*}{\makecell[c]{3D-based\\ Methods}}&\multirow{5}{*}{Voxel Grids}&3DShapeNet~\cite{2015-CVPR-3DShapeNet}&2015&Convolutional BDN&24$\times$24$\times$24&Contrastive Divergence&&\checkmark\\\cline{3-9}
		&&VolumetricCNN~\cite{2016-CVPR-VolumetricCNN}&2016& CNN&512&Classification&\checkmark&\\\cline{3-9}
		&&3DMatch~\cite{2017-CVPR-3DMatch}&2017&3D ConvNet&512&Contrastive&&\checkmark\\\cline{3-9}
		&&3DSmoothNet~\cite{2019-CVPR-3DSmoothNet}&2019&
		CNN&16&Batch Hard&&\checkmark\\\cline{3-9}
		&&SpinNet~\cite{2021-CVPR-SpinNet,2023-TPAMI-YOTO}&2021&Transformer+3DCCN&32&Contrastive&\checkmark&\checkmark\\\cline{2-9}
		&\multirow{10}{*}{Raw 3D Points}&PointNet~\cite{2017-CVPR-PointNet}&2017&CNN&1024&Regularization Softmax&&\checkmark\\\cline{3-9}
		&&PointNet++~\cite{2017-ANIPS-Pointnet++}&2017&PointNet&&Cross Entropy&\checkmark&\checkmark\\\cline{3-9}
		&&CGF~\cite{2017-ICCV-CGF}&2017&DNN&32&Triplet&&\checkmark\\\cline{3-9}
		&&PPFNet~\cite{2018-CVPR-PPFNet}&2018&PointNet&64&N-tuple&&\checkmark\\\cline{3-9}
		&&PPF-FoldNet~\cite{2018-ECCV-PPF-FoldNet}&2018&MLP&512&N-tuple&&\checkmark\\\cline{3-9}
		&&DeepVcp~\cite{2019-ICCV-DeepVcp}&2019&PointNet++&32&L1+L2&\checkmark&\checkmark\\\cline{3-9}
		&&RelativeNet~\cite{2019-CVPR-RelativeNet}&2019&PPF-FoldNet&&Chamfer&\checkmark&\\\cline{3-9}
		&&L3Ds~\cite{2022-RAL-L3Ds}&2022&TNet+PointNet&32&Contrastive&&\checkmark\\\cline{3-9}
		&&Poiesi \etal~\cite{2022-TPAMI-LGAD}&2022&QNet+PointNet++&32&Hardest-contrastive&&\checkmark\\\cline{3-9}
		&&LEAD~\cite{2022-TPAMI-ULOL}&2022&Spherical CNN&512&Chamfer Distance&\checkmark&\\\hline
		\multicolumn{2}{|c|}{\multirow{6}{*}{2D-based Methods}}&LORAX~\cite{2017-CVPR-LORAX}&2017&DNN&1032&Pixel-wise Error+ICP&&\checkmark\\\cline{3-9}
		\multicolumn{2}{|c|}{}&MVDesc~\cite{2018-ECCV-LMMD}&2018&MatchNet&32&Double-margin Contrastive&&\\\cline{3-9}
		\multicolumn{2}{|c|}{}&Li \etal~\cite{2020-CVPR-ELLM}&2020&CNN&32&Batch-hard triplet&\checkmark&\checkmark\\\cline{3-9}
		\multicolumn{2}{|c|}{}&Gojcic \etal~\cite{2020-CVPR-LM3P}&2020&FCGF&32&Hardest
		Contrastive&\checkmark&\checkmark\\\cline{3-9}
		\multicolumn{2}{|c|}{}&DeLightLCD~\cite{2022-IEEESensors-DelightLCD}&2022&DNN&1$\times$300$\times$32&Binary Cross Entropy&\checkmark&\\\hline
	\end{tabular}
	\begin{tablenotes}
		\item \textbf{Size} and \textbf{EtE} denotes the descriptor size and end-to-end learning, respectively.
	\end{tablenotes}
	\label{tab:sum-hand-local-descriptor-methods}
\end{table*}
\section{LPR Techniques: Local Descriptor}\label{sec:lpr-techniques-local-descriptor}
The local descriptor is a compact representation of regions or points, capturing distinctive characteristics such as texture, color, density, or shape. Local descriptor-based methods typically extract keypoints and employ local descriptors to characterize their surrounding context. They generally fall into either 3D-based or 2D-based categories based on the nature of descriptors. Table \ref{tab:sum-hand-local-descriptor-methods} contains a systematic summary. 
\subsection{Handcrafted Methods}\label{sec:ld-handcrafted-methods}
\subsubsection{3D-based Methods}\label{sec:ld-handcrafted-methods-3d-based-methods}
We roughly categorized handcrafted 3D local descriptors into two groups based on a Local Reference Frame (LRF).

\textbf{LRF-based Methods}. LRF rigidly transforms the patch into canonical representation by selecting neighborhood points to build a covariance matrix and computing the eigenvector as reference axes. While initially designed for PCR, it is also applicable to place recognition. Several methods focus on encoding geometric information, such as normal~\cite{2019-RAL-ISHOT}, height~\cite{2020-IS-AEAC}, and mesh~\cite{2013-IJCV-ROPS}, within the LRF to achieve precise geometric descriptions. Others enhance the stability of LRF using weighted projection vectors~\cite{2017-PR-TOLDI} or sign disambiguation~\cite{2010-ACM-USC}.

\textbf{LRF-free Methods}. Other methods ensure rotation invariance by avoiding LRF construction and focusing solely on the underlying geometry of the local surface. Early methods directly count surrounding geometric information, such as height~\cite{2013-ICRA-3DGestalt}, distance~\cite{2017-IROS-GLAROT-3D}, and density~\cite{1999-TPAMI-SpinImage}. Subsequently, several methods encode point distribution~\cite{2020-PR-HOPPF} and sparse triangulated landmark~\cite{2016-ICRA-NBLD}.
\subsubsection{2D-based Methods}\label{sec:2d-based-methods}
These methods build handcrafted local descriptors from the projected 2D image, followed by an image matching problem. Spherical view and Bird-Eye-View (BEV) are two representative projection approaches. 

\textbf{Spherical View}. Projecting point clouds into a spherical or range image effectively mitigate orientation ambiguity. Steder \etal~\cite{2010-ICRA-RPRF} pioneer the projection of point clouds to range images for place recognition. They extract the local descriptor vector~\cite{2009-IROS-ROMO} and evaluated candidate transformations through keypoint reprojection. Afterward, several works extend the method~\cite{2010-ICRA-RPRF} using Normal Aligned Radial Features (NARF)~\cite{2011-IROS-NARF}, Speeded Up Robust Features (SURF)~\cite{2013-TIM-3SMA} and ORB~\cite{2018-IEEESensors-RPCA, 2021-ICRA-AIL}.

\textbf{BEV}. Several works incorporate proposal-wise features from BEV images into image matching. BVMatch~\cite{2021-RAL-BVMatch} extract the maximum index map of the Log-Gabor filter responses, employing BEV feature transform and BoW for place recognition. It provides relative poses and effectively overcomes sparsity and intensity distortion. Luo \etal~\cite{2022-TIV-HOPN} apply FAST~\cite{2016-ECCV-FAST} detectors on the BEV image and construct a global descriptor using 3D normals. It showcases superior localization capability in large-scale scenarios.
\subsection{Learning-based Methods}\label{sec:ld-learning-based-methods}
\subsubsection{3D-based Methods}
Learning-based 3D local descriptors typically employ 3D CNNs to encode point cloud patches, divided into voxel grids and raw 3D points according to network inputs.

\textbf{Voxel Grids}. The pioneering work 3DMatch~\cite{2017-CVPR-3DMatch} transform patches into voxel grids of Truncated Signed Distance Function (TSDF) and employed eight convolutional layers to learn the descriptor. Several works extend this idea to more informative encoding manner such as binary occupancy~\cite{2016-CVPR-VolumetricCNN}, multi-label occupancy~\cite{2015-CVPR-3DShapeNet}, smoothed density value~\cite{2019-CVPR-3DSmoothNet}, and spherical voxelization~\cite{2021-CVPR-SpinNet,2023-TPAMI-YOTO}.

\textbf{Raw 3D Points.} An alternative method involves direct processing of the raw point cloud data.

(a) PointNet Family. PointNet~\cite{2017-CVPR-PointNet} is a pioneering work of learning from unordered point clouds, which learns a symmetry function approximated by a Multi-Layer Perceptron (MLP) to handle detailed shapes. Subsequently, several works extend PointNet~\cite{2017-CVPR-PointNet} by ball query search~\cite{2017-ANIPS-Pointnet++}, normals~\cite{2018-CVPR-PPFNet}, FoldingNet decoder~\cite{2018-ECCV-PPF-FoldNet}, orientation~\cite{2019-CVPR-RelativeNet}, LRF~\cite{2022-RAL-L3Ds,2022-TPAMI-LGAD}, and semantics~\cite{2019-ICCV-DeepVcp}.

(b) Other Methods. Compact Geometric Features (CGF)~\cite{2017-ICCV-CGF} trains a deep network to map from the high-dimensional space of spherical histograms to a low-dimensional Euclidean space. LEAD~\cite{2022-TPAMI-ULOL} combines spherical CNNs to learn the equivariant representation.
\subsubsection{2D-based Methods}
Several works infer local descriptors using well-established 2D CNNs from projected 2D images. They demonstrate superior performance in the task of 3D shape recognition and retrieval. LORAX~\cite{2017-CVPR-LORAX} and DeLightLCD~\cite{2022-IEEESensors-DelightLCD} employ a Deep Neural Network (DNN) auto-encoder and attention to enhance descriptor descriptiveness on 2D depth images, respectively. Another spectrum of research fuses multi-view features into descriptors by soft-view pooling~\cite{2020-CVPR-ELLM}, graphical model~\cite{2018-ECCV-LMMD}, and spectral relaxation~\cite{2020-CVPR-LM3P}.
\subsection{Observations and Implications}\label{sec:ld-summary}
This section review some representative methods and more comprehensive surveys of local descriptors are available in~\cite{2014-TPAMI-Survey-Local-3ORI}. Although local descriptors find wide applications in tasks such as registration and object recognition, they are not the preferred methods for PR. There are mainly the following reasons:

\textbf{(i)} Viewpoint changes can affect the accuracy of 3D keypoints, rendering them unsuitable for matching. Moreover, they may not effectively handle data noise and object occlusions. \textbf{(ii)} The usage of 3D local descriptors~\cite{2013-ICRA-3DGestalt,2016-ICRA-NBLD} can be challenging as it requires dense point clouds, which is computationally expensive and may not work well with sensors like Velodyne VLP-16~\cite{2020-ICRA-OneShot} that produce sparse point clouds. \textbf{(iii)} While converting point clouds into images~\cite{2010-ICRA-RPRF,2021-RAL-BVMatch,2022-IEEESensors-DelightLCD} can use mature image processing techniques, this results in the loss of geometric information, making it unsuitable for large-scale scenarios. 
\begin{table*}[!t]
	\tiny
	\centering
	\caption{A summary of global descriptor-based methods.}
	\begin{tabular}{|m{1.05cm}<{\centering}|m{0.7cm}<{\centering}|r|m{0.3cm}<{\centering}|c|m{1.06cm}<{\centering}|m{0.65cm}<{\centering}|c|m{0.35cm}<{\centering}|}
		\hline
		\rowcolor{gray!20} 
		\multicolumn{9}{|c|}{\textbf{Handcrafted Methods}}\\\hline
		\multicolumn{3}{|c|}{\textbf{Methods}} &\textbf{Year}&	\multicolumn{2}{c|}{\textbf{Metric}}&\textbf{Size}&\textbf{Feature}&\textbf{Code}\\\hline
		\multirow{10}{*}{BEV}&\multirow{6}{*}{\makecell[c]{SC\\ Family}}&SC~\cite{2018-IROS-ScanContext}&2018 &\multicolumn{2}{c|}{L0 Norm+Cosine Distance}&60$\times$20 &Height&\checkmark\\\cline{3-9}
		&&ISC~\cite{2020-ICRA-ISC}&2020&\multicolumn{2}{c|}{Cosine Distance}&60$\times$20&Intensity&\checkmark\\\cline{3-9}
		&&SC++~\cite{2021-TRO-ScanContext++}&2021&\multicolumn{2}{c|}{L1 Norm+Cosine Distance}&60$\times$20&Height&\checkmark\\\cline{3-9}
		&&FreSCo~\cite{2022-ICRACV-FreSCo}&2022&\multicolumn{2}{c|}{L1 Norm+Cosine Distance}&20$\times$120&Height&\checkmark\\\cline{3-9}		
		&&FSC~\cite{2022-IEEESensors-FSC}&2022&\multicolumn{2}{c|}{F-Norm}&&Height+Intensity&\\\cline{3-9}
		&&Ou \etal~\cite{2023-RAL-PROL}&2023&\multicolumn{2}{c|}{Cosine Distance}&5$\times$20$\times$60&Density+Height&\\\cline{2-9}	
		&\multirow{4}{*}{\makecell[c]{Pairwise\\ Matching}}&LiDAR Iris ~\cite{2020-IROS-LiDAR-Iris}&2020&\multicolumn{2}{c|}{Hamming Distance}&80$\times$360&Height&\checkmark\\\cline{3-9}
		&&RING~\cite{2022-IROS-RING}&2022&\multicolumn{2}{c|}{Circular Cross-correlation}&120$\times$120&Occupancy&\checkmark\\\cline{3-9}
		&&RING++~\cite{2023-TRO-RING++}&2022&\multicolumn{2}{c|}{Circular Cross-correlation}&120$\times$120&Height+Occupancy&\checkmark\\\cline{3-9}
		&&PGHCI~\cite{2022-ISPRSJ-PGHCI}&2022&\multicolumn{2}{c|}{JS Divergence+Pixel Values}&40$\times$20&Height&\\\hline
		\multirow{3}{*}{Discretization}&\multirow{3}{*}{\makecell[c]{Fixed-\\size}}&Magnusson \etal~\cite{2009-ICRA-NDT-ALD}&2009 &\multicolumn{2}{c|}{Euclidean Distance}&&Shape&\\\cline{3-9}
		&&Lin \etal~\cite{2019-arXiv-Fast-Complete}&2019&\multicolumn{2}{c|}{Normalized Cross-correlation}&60$\times$60&Shape&\checkmark\\\cline{3-9}
		&&Cao \etal~\cite{2021-TIE-SIV}&2021&\multicolumn{2}{c|}{Euclidean Distance}&360$\times$180&Context+Layout&\\\cline{2-9}
		&\multirow{2}{*}{\makecell[c]{Unfixed-\\size}}&DELIGHT~\cite{2018-ICRA-Delight}&2018&\multicolumn{2}{c|}{Chi-squared Test}&256&Intensity&\\\cline{3-9}
		&&Mo \etal~\cite{2020-IROS-AFR}&2020&\multicolumn{2}{c|}{L2-Norm+Chi-square Test}&&Density+Intensity+Height&\checkmark\\\hline
		\multicolumn{2}{|c|}{\multirow{4}{*}{Point}}&Z-Projection~\cite{2011-SSRR-Z-Projection}&2011 &\multicolumn{2}{c|}{$\chi^2$ Distance+Sørensen Distance}&101&Normal&\\\cline{3-9}
		\multicolumn{2}{|c|}{}&Fast Histogram~\cite{2015-IROS-FastHistogram}&2015&\multicolumn{2}{c|}{Wasserstein Metric}&80&Height&\\\cline{3-9}
		\multicolumn{2}{|c|}{}&M2DP~\cite{2016-IROS-M2DP}&2016&\multicolumn{2}{c|}{L2-Norm}&192&Density&\checkmark\\\cline{3-9}
		\multicolumn{2}{|c|}{}&C-M2DP~\cite{2019-CASE-CM2DP}&2019&\multicolumn{2}{c|}{L2 Distance}&576&Color+Shape&\\\hline
		\rowcolor{gray!20} \multicolumn{9}{|c|}{\textbf{Learning-based Methods}}\\\hline
		\multicolumn{3}{|c|}{\textbf{Methods}}&\textbf{Year}&\textbf{Backbone}&\textbf{Aggregator}&\textbf{Size}&\textbf{Loss}&\textbf{EtE}\\\hline
		\multirow{11}{*}{\makecell[c]{Point} }&\multirow{4}{*}{\makecell[c]{Pointwise\\ MLP}}&Pointnetvlad~\cite{2018-CVPR-Pointnetvlad}&2018&PointNet&NetVLAD&256&Lazy triplet and quadruplet&\checkmark\\\cline{3-9}
		&&PCAN~\cite{2019-CVPR-PCAN}&2019&PointNet&NetVLAD&256&Lazy quadruplet&\checkmark\\\cline{3-9}		
		&&SOE-Net~\cite{2021-CVPR-SOENet}&2021&PointOE&NetVLAD&256&HPHN quadruplet&\checkmark\\\cline{3-9}		
		&&LCD-Net~\cite{2022-TRO-LCDNet}&2022&PV-RCNN&NetVLAD&256&Triplet&\checkmark\\\cline{2-9}
		&\multirow{2}{*}{\makecell[c]{Point\\ Conv}}&DH3D~\cite{2020-ECCV-DH3d}&2020&FlexConv+SE block&NetVLAD&256&N-tuple&\checkmark\\\cline{3-9}
		&&EPC-Net~\cite{2022-TIP-EPC-Net}&2022&PPCNN&VLAD&256&Lazy quadruplet&\checkmark\\\cline{2-9}
		&\multirow{5}{*}{\makecell[c]{Graph}}&LPD-Net~\cite{2019-ICCV-LpdNet}&2019&PointNet&NetVLAD&256&Lazy quadruplet&\checkmark \\\cline{3-9}	
		&&DAGC~\cite{2020-ICMR-DAGC}&2020&ResGCN&NetVLAD&256&Lazy quadruplet&\\\cline{3-9}
		&&SR-Net~\cite{2020-CGF-SR-Net}&2020&SGC+SAM&NetVLAD&1024&Lazy quadruplet&\\\cline{3-9}
		&&vLPD-Net~\cite{2021-IROS-vLPD-Net}&2021&LPD-Net+S-ARN&MinkPool&&Joint loss&\\\cline{3-9}		
		&&PPT-Net~\cite{2021-ICCV-PPT-Net}&2021&Transformer&VLAD&256&Lazy quadruplet&\checkmark\\\hline
		\multirow{12}{*}{\makecell[c]{Discretization}}&\multirow{8}{*}{\makecell[c]{Sparse\\Repre}}&MinkLoc3D~\cite{2021-WACV-Minkloc3d}&2021&FPN&GeM&256&Triplet margin&\checkmark\\\cline{3-9}	
		&&MinkLoc++~\cite{2021-IJCNN-MinkLoc++}&2021&ResNet18+FPN&GeM&256&Triplet margin&\checkmark\\\cline{3-9}
		&&EgoNN~\cite{2021-RAL-EgoNN}&2021&CNN&GeM&256&Triplet margin&\checkmark\\\cline{3-9}		
		&&TransLoc3D~\cite{2021-arXiv-Transloc3d}&2021&Transformer&NetVLAD&256&Triplet margin&\checkmark\\\cline{3-9}		
		&&MinkLoc3Dv2~\cite{2022-ICPR-MinkLoc3Dv2}&2022&FPN&GeM&256&Modified Smooth-AP&\checkmark\\\cline{3-9}	
		&&MinkLoc3D-SI~\cite{2022-RAL-MinkLoc3D-SI}&2022&FPN&GeM&256&Triplet margin&\checkmark\\\cline{3-9}		
		&&SVTNet~\cite{2022-AAAI-SVTNet}&2022&Transformer&GeM&256&Triple&\checkmark\\\cline{3-9}		
		&&LoGG3D-Net~\cite{2022-ICRA-LoGG3D-Net}&2022&U-Net&O2P+ePN&256&Contrastive+Quadruplet&\checkmark\\\cline{2-9}		
		&\multirow{4}{*}{\makecell[c]{Dense\\Repre}}&SpoxelNet~\cite{2020-IROS-SpoxelNet}&2020&CNN&NetVLAD&&Lazy quadruplet&\checkmark\\\cline{3-9}
		&&VBRL~\cite{2020-IROS-VRLF}&2020&&&&Modality norm&\\\cline{3-9}
		&&HiTPR~\cite{2022-ICRA-HiTPR}&2022&Transformer&Max pooling&1024&Lazy quadruplet&\\\cline{3-9}			
		&&NDT-Transformer~\cite{2021-ICRA-NDT-Transformer}&2022&Transformer&NetVLAD&256&Lazy quadruplet&\checkmark\\\hline	
		\multirow{8}{*}{\makecell[c]{Projection}}&\multirow{6}{*}{\makecell[c]{Spherical\\View}}&Yin \etal~\cite{2017-ROBIO-E3LB}&2017&DNN&&&Contrastive&\\\cline{3-9}
		&&MMCS-Net~\cite{2022-TIE-MMCS-Net}&2022&Siamese CNNs&NetVLAD&&Contrastive&\\\cline{3-9}
		&&SeqOT~\cite{2022-TIE-SeqOT}&2022&Transformer&GeM&256&Triplet&\checkmark\\\cline{3-9}
		&&OverlapTransformer~\cite{2022-RAL-OverlapTransformer}&2022&Transformer&NetVLAD&256&Lazy triplet&\checkmark\\\cline{3-9}			
		&&AttDLNet~\cite{2021-arXiv-AttDLNet}&2021&DarkNet53&Max pooling&1024&Cosine similarity&\checkmark\\\cline{3-9}	
		&&OREOS~\cite{2019-IROS-OREOS}&2019&CNN&&&Triplet&\\\cline{2-9}
		&\multirow{2}{*}{BEV}&SCI~\cite{2019-RAL-SCI}&2019&LeNet&&&Categorical cross-entropy&\checkmark\\\cline{3-9}
		&&DiSCO~\cite{2021-RAL-DiSCO}&2021&U-Net&&1024&Quadruplet+KL divergence&\checkmark\\\hline
	\end{tabular}
	\begin{tablenotes}
	\item \textbf{Size} and \textbf{EtE} denotes the descriptor size and end-to-end learning, respectively. \textbf{Sparse/dense repre} denotes sparse/dense representation. \textbf{Point Conv} means point convolution.
	\end{tablenotes}
	\label{tab:sum-global-descriptor}
\end{table*}
\section{LPR Techniques: Global Descriptor}\label{sec:lpr-techniques-global-descriptor}
The global descriptor captures the overall features of a scene, providing a holistic view of the data rather than focusing on specific regions or points. Table \ref{tab:sum-global-descriptor} contains a systematic summary of global descriptor-based methods.
\subsection{Handcrafted Methods}\label{sec:gl-handcrafted-methods}
\subsubsection{BEV-based Methods}
BEV projection gains significant attention in the robotics community due to its ability to enhance algorithm efficiency through dimension reduction, making it highly suitable for real-time applications. Scan Context (SC)~\cite{2018-IROS-ScanContext} family and pairwise matching are two mainstream methods. 

\textbf{SC Family}. The pioneering work SC~\cite{2018-IROS-ScanContext} partitions the horizontal space into discrete bins while maintaining the points' maximum height to generate a 2D matrix descriptor. It utilizes the ring key to search for potential matches and conducts a column-wise comparison to identify the closest one. This method demonstrates promising performance but may fail when dealing with significant lateral offsets. Subsequently, researchers propose a series of SC-based variant methods, which employ the polar and cart context~\cite{2021-TRO-ScanContext++}, intensities~\cite{2020-ICRA-ISC}, frequency domain~\cite{2022-ICRACV-FreSCo}, F-norm~\cite{2022-IEEESensors-FSC}, and Spatial Binary Pattern (SBP)~\cite{2023-RAL-PROL} to enhance performance.

\textbf{Pairwise Matching}. LiDAR Iris~\cite{2020-IROS-LiDAR-Iris} draws inspiration from human iris signatures, utilizing LoG-Gabor filtering and thresholding to create binary signature images, then measuring descriptor similarities using Hamming distance. Some methods also employ weighted distances~\cite{2022-ISPRSJ-PGHCI} and orientation-invariant metrics~\cite{2022-IROS-RING,2023-TRO-RING++} for pairwise similarity computation.
\subsubsection{Discretization-based Methods}
The discretization processing transforms the point cloud into 3D discrete representations, categorized into fixed and unfixed size-based approaches.

\textbf{Fixed-size Discretization}. Magnusson \etal~\cite{2009-ICRA-NDT-ALD} maps the point cloud to Normal Distribution Transform (NDT) voxels and creates a histogram based on the probability density function of the local surface. They calculate the descriptor similarities using weighted Euclidean distances. This method demonstrates the potential of NDT descriptors for place recognition. Subsequently, researchers introduce techniques like k-means++ clustering~\cite{2017-IROS-S3PC} and normalized cross-correlation metrics~\cite{2019-arXiv-Fast-Complete} to enhance generalization. Cao \etal~\cite{2021-TIE-SIV} also generate a numerical descriptor by detecting contours and computing spectrum energies from wedge-shaped voxels.

\textbf{Unfixed-size Discretization}. DELIGHT~\cite{2018-ICRA-Delight,2020-IROS-AFR} divides the support region into two concentric spheres and gets non-overlapping bins by horizontal and azimuthal divisions. It computes intensity histograms for each bin and assesses descriptor similarities by chi-squared tests. Notably, the descriptor can be local or global based on the descriptor's radius and center point. 
\subsubsection{Point-based Methods}
Several methods treat place recognition as a histogram matching problem, encoding angle and height information and calculating histogram similarities using Wasserstein metric~\cite{2015-IROS-FastHistogram}, Sørensen~\cite{2011-SSRR-Z-Projection}, and $\chi^2$ distances~\cite{2011-SSRR-Z-Projection}. They achieve rotation invariance and overcome noises. A parallel track of works follow a projection-based scheme. M2DP~\cite{2016-IROS-M2DP} and C-M2DP~\cite{2019-CASE-CM2DP} project the point cloud onto 2D planes using azimuth and elevation angles, counting point densities to create the descriptor. Multi-view density signatures enable accurate descriptions with fewer computational resources, making it particularly effective for sparse point clouds.
\subsection{Learning-based Methods}\label{sec:gl-learning-based-methods}
\subsubsection{Point-based Methods}
One prevalent approach directly utilizes the inherent 3D spatial information for LiDAR point cloud processing, involving point-wise MLP, point convolution, and graph representation.

\textbf{Pointwise MLP}. The pioneering work PointNetVLAD~\cite{2018-CVPR-Pointnetvlad} combines PointNet~\cite{2017-CVPR-PointNet} for local feature extraction and NetVLAD~\cite{2016-CVPR-NetVLAD} for global descriptor generation. It employs metric learning and introduces the lazy triplet and quadruplet loss functions to enhance generality. Afterward, PCAN~\cite{2019-CVPR-PCAN} and SOE-Net~\cite{2021-CVPR-SOENet} improve high-dimensional feature representation by incorporating attention mechanisms. LCD-Net~\cite{2022-TRO-LCDNet} utilizes the PointVoxel-RCNN (PV-RCNN)~\cite{2020-CVPR-PVRCNN} architecture and combines the feature extraction capabilities of DNN with transport theory algorithms.

\textbf{Point Convolution}. DH3D~\cite{2020-ECCV-DH3d} introduces a Siamese network for local feature detection, description, and global descriptor extraction in a single forward pass. It incorporates multi-level spatial contextual information and channel-wise feature correlations. EPC-Net~\cite{2022-TIP-EPC-Net} is a compact model based on edge convolution, simplifying the process with spatial-adjacent matrices and proxy points. It achieves excellent performance while significantly reducing computational memory.

\textbf{Graph Representation}. Graph networks efficiently captures underlying geometric and shape properties, allowing for feature comparison across multiple locations within graphs. LPD-Net~\cite{2019-ICCV-LpdNet} and vLPD-Net~\cite{2021-IROS-vLPD-Net} extract multiple local features, including curvature, height, and density, and employ a Graph Neural Network (GNN) for feature aggregation. Several works integrate an attention module to discern task-relevant features~\cite{2020-ICMR-DAGC}, learn spatial relationships between regions~\cite{2021-ICCV-PPT-Net}, and mitigated the influence of movable noises~\cite{2020-CGF-SR-Net}.
\subsubsection{Discretization-based Methods}
We divide discretization-based methods into sparse and dense representations based on voxel density.

\textbf{Sparse Representation}. Three approaches for generating global descriptors using 3D CNNs on sparse volume representations are Feature Pyramid Network (FPN), transformer network, and other methods.

(a) FPN. MinkLoc3D~\cite{2021-WACV-Minkloc3d}, a pioneering work based on sparse voxelization, employs a 3D FPN~\cite{2017-CVPR-FPN} and Generalized-Mean (GeM)~\cite{2017-CVPR-GEM} pooling for global descriptor generation. It showcases a simple and elegant architecture, highlighting the potential of sparse voxelized representation and sparse convolutions for efficient 3D feature extraction. Later works extend enhance recognition performance by incorporating intensity~\cite{2022-RAL-MinkLoc3D-SI}, image~\cite{2021-IJCNN-MinkLoc++}, and attention~\cite{2021-RAL-EgoNN,2022-ICPR-MinkLoc3Dv2}.

(b) Transformer Network. TransLoc3D~\cite{2021-arXiv-Transloc3d} re-weights features from multiple receptive scales using an attention map and incorporates external attention layers for capturing long-range contextual information. SVT-Net~\cite{2022-AAAI-SVTNet}, introduces two types of transformers to capture short-range local features and long-range contextual features, respectively. Despite a shallow network architecture, it generates descriptive descriptors.

(c) Other Methods. LoGG3D-Net~\cite{2022-ICRA-LoGG3D-Net} utilizes the sparse point-voxel convolution for high-dimensional feature embedding and introduces a local consistency loss for feature similarity maximization. It exhibits superior end-to-end performance, operating in near real-time.

\textbf{Dense Representation}. Likewise, we broadly categorize dense representation-based approaches into transformer network, DNN, and other methods.

(a) Transformer Network. HiTPR~\cite{2022-ICRA-HiTPR} utilizes a short-range transformer to extract local features within cells and a long-range transformer to encode global relations among the cells. It enhances the relevance of local neighbors and global contextual dependencies. NDT-Transformer~\cite{2021-ICRA-NDT-Transformer} introduces a novel network with three stacked transformer encoders, learning a global descriptor from discrete NDT cells. It is a valuable addition to NDT-based SLAM and MCL methods.

(b) DNN. SpoxelNet~\cite{2020-IROS-SpoxelNet} voxelizes the point cloud in spherical coordinates, representing voxel occupancy using ternary values. It extracts multi-scale structural features and generates a global descriptor by concatenating features from various directions. This method effectively handles occlusion and moving objects in crowded indoor spaces.

(c) Other Methods. Voxel-based Representation Learning (VBRL)~\cite{2020-IROS-VRLF} tackles long-term place recognition by jointly learning voxel importance and feature modalities using structured sparsity-inducing norms. It integrates all features into a unified regularized optimization formulation.
\subsubsection{Classification-based Methods}
Several approaches address the place recognition problem using classifiers. FastLCD~\cite{2021-JAG-FastLCD} encodes multi-modality features into a global descriptor, detecting candidate loop closures using supervised learning and rejecting false positives through cross-validation and post-verification. Habich \etal~\cite{2021-AIM-HIBH} perform loop searches within a variable radius based on the eigenvalue of the position covariance matrix and predict loops using a classifier. 
\subsubsection{Projection-based Methods}
Following the taxonomy of learning-based 2D local descriptors (Section \ref{sec:2d-based-methods}), this section also categorizes projection methods into spherical view and BEV.

\textbf{Spherical View}. Using spherical projection images as input, three representative methods include siamese network, transformer network, and 2D CNN only.

(a) Siamese Network. 
Yin \etal~\cite{2017-ROBIO-E3LB} transform the point cloud into a one-channel image and use a Siamese CNN to convert LCD into a similarity modeling problem, improving search efficiency by combining Euclidean metric and kd-tree. MMCS-Net~\cite{2022-TIE-MMCS-Net} incorporates a Siamese CNN with shallow-deep feature fusion and a cascaded attention mechanism to handle pseudo images. It effectively strikes a favorable balance between effectiveness and efficiency.
\begin{figure}[tbp]
	\centering
	\subfigure[Spherical View]{\includegraphics[height=1.9cm] {image/2022-RAL-OverlapTransformer-1.jpg}}
	\subfigure[BEV]{\includegraphics[height=1.9cm] {image/2021-RAL-DiSCO-3.jpg}}
	\caption{Two projection-based methods. (a) and (b) are originally shown in ~\cite{2022-RAL-OverlapTransformer} and ~\cite{2021-RAL-DiSCO}, respectively.}\label{fig:data-driven-projection}
\end{figure}

(b) Transformer Network. SeqOT~\cite{2022-TIE-SeqOT} employs multi-scale transformers to generate sub-descriptors that fuse spatial and temporal information from sequential LiDAR range images. It ensures robustness to viewpoint changes and scan order, enabling reliable place recognition even in opposite directions. As depicted in Figure \ref{fig:data-driven-projection}(a), OverlapTransformer~\cite{2022-RAL-OverlapTransformer} extracts features from range images and integrates a transformer to capture relative feature locations, demonstrating fast running speed and robust generalization. AttDLNet~\cite{2021-arXiv-AttDLNet} incorporates a four-layer attention network to capture long-range context and inter-feature relationships. 

(c) 2D CNN Only. OREOS~\cite{2019-IROS-OREOS} uses 2D convolutional and max pooling layers to extract features from 2D range images, enhancing the descriptor performance via a triple loss function and strong negative mining strategy. It efficiently computes the descriptor while enabling long-term 3-DoF metric localization in outdoor environments.

\textbf{BEV.} Two representative BEV-based methods are encoder-decoder network and 2D CNN only.

(a) Encoder-Decoder Network. As shown in Figure \ref{fig:data-driven-projection}(b), DiSCO~\cite{2021-RAL-DiSCO} employs an encoder-decoder network to extract descriptors and estimates relative orientation through Fourier-Mellin Transform and differentiable phase correlation. It enhances the interpretability and efficiency of the feature extractor.

(b) 2D CNN Only. Scan Context Image (SCI)~\cite{2019-RAL-SCI} extends SC~\cite{2018-IROS-ScanContext} into three channels, enabling robot localization on a grid map through a convolutional neural network-based place classification. It demonstrates robust year-round localization with only a single day of learning.
\subsection{Observations and Implications}
Global descriptors are currently the most popular place recognition method, which can provide information about the entire scene, unaffected by local changes. The progress of deep learning in 3D computer vision paves the way for data-driven methods in LPR. Several observations are summarized as follows:

For the handcrafted part: \textbf{(i)} BEV~\cite{2018-IROS-ScanContext,2021-TRO-ScanContext++} demonstrates superior performance in flat structural environments but may yield poor results when the LiDAR's z-axis changes, as these methods assume local planar vehicle motion. \textbf{(ii)} Discretization-based methods~\cite{2009-ICRA-NDT-ALD,2019-arXiv-Fast-Complete,2021-TIE-SIV} can describe the local surface using robust mathematical theories. However, increasing the resolution will significantly incur a heavy computational burden. \textbf{(iii)} Point-based methods~\cite{2011-SSRR-Z-Projection,2015-IROS-FastHistogram,2016-IROS-M2DP} are the most basic global descriptor methods. However, they require expensive neighbor searching to establish topological relationships. Furthermore, projection operations may result in information loss and cause potential false positives.

For the learning part: \textbf{(i)} Learning-based methods are efficient and accurate but require large, clean datasets and often necessitate transfer learning to address real-world issues like noise and occlusions. \textbf{(ii)} Transformers ~\cite{2022-ICRA-HiTPR,2021-ICRA-NDT-Transformer} excel at capturing contextual relationships, enabling reliable recognition in cluttered environments. However, their substantial computational demands constrain the batch size for metric learning. Sparse convolutional architectures~\cite{2021-WACV-Minkloc3d,2022-RAL-MinkLoc3D-SI} excel at generating informative local features while struggling to discriminate feature size in dynamic scenarios. \textbf{(iii)} 
Point-based methods~\cite{2018-CVPR-Pointnetvlad,2019-CVPR-PCAN} handle unordered data well but may miss local spatial details. Classification-based methods~\cite{2021-JAG-FastLCD,2021-AIM-HIBH} assign higher weights to informative features during training. However, the specific contribution of each weak classifier to the overall prediction may be less interpretable. Projection-based methods~\cite{2022-TIE-MMCS-Net,2019-IROS-OREOS} are efficient and interpretable but may lose information due to dimensionality reduction.
\section{LPR Techniques: Segments}\label{sec:lpr-techniques-segments}
Segments are meaningful region divisions characterized by similar geometric properties. These methods divide the point cloud into segments and three typical methods are shown in Figure \ref{fig:segments}.
\begin{figure*}[tbp]
	\centering
	\subfigure[Matching]{\includegraphics[width=3.5cm] {image/SegMatch-v2.jpg}}
	\hspace{0.1cm}
	\subfigure[Classification]{\includegraphics[width=3.7cm] {image/SegMap-v2.jpg}}
	\hspace{0.1cm}
	\subfigure[Similarity]{\includegraphics[width=4.6cm] {image/Locus-v4.jpg}}
	\caption{An illustration of three representative segment-based methods. (a)-(c) are originally shown in ~\cite{2017-ICRA-Segmatch}, ~\cite{2018-RSS-SegMap}, and ~\cite{2021-ICRA-Locus}, respectively.}\label{fig:segments}
\end{figure*}
\subsection{Handcrafted Methods}\label{sec:segment-handcrafted-methods}
\subsubsection{Matching-based Methods}
Segment-based matching for finding correspondences primarily comprises the SegMatch~\cite{2017-ICRA-Segmatch} family and other methods.

\textbf{SegMatch Family}. The pioneering work SegMatch~\cite{2017-ICRA-Segmatch}, depicted in Figure \ref{fig:segments}(a), employs Euclidean clustering to partition the point cloud into segments and extracts eigenvalue-based features. It effectively identifies potential correspondences using random forest and Random Sample Consensus (RANSAC)~\cite{1981-CACM-RANSAC}-based geometric verification. Dubé \etal~\cite{2018-RAL-ISBL} enhance SegMatch~\cite{2017-ICRA-Segmatch} by tracking a single segment using region-growing-based incremental segmentation. Moreover, certain studies effectively integrate SegMatch~\cite{2017-ICRA-Segmatch} into traditional LiDAR SLAM~\cite{2019-ICMA-LLOAM} and multi-robot systems~\cite{2017-IROS-AOMS}.

\textbf{Other Methods}. RDC-SLAM~\cite{2021-TITS-RDC-SLAM} combines an eigenvalue-based segment descriptor, K Nearest Neighbors (KNN) search, and RANSAC-based verification~\cite{1981-CACM-RANSAC} to refine relative poses. Gong \etal~\cite{2021-PR-ATFF} construct a spatial relation graph to represent segments, effectively capturing general spatial relations between irregular clusters.
\subsubsection{Similarity-based Methods}
Seed~\cite{2020-IROS-Seed} develops a segmentation-based egocentric descriptor, incorporating topological information into SC-based place recognition~\cite{2018-IROS-ScanContext}. It achieves translation and rotation invariance by utilizing the inner topological structure of segmented objects. 
\subsection{Learning-based Methods}\label{sec:segment-learning-based-methods}
\subsubsection{Matching-based Methods}
Tinchev \etal~\cite{2018-IROS-STWF} encode geometric properties and point distribution of segments to extract repeatable oriented key poses, which are matched using reliable shape descriptors and a Random Forest. However, significant changes in the sensor's vantage point could negatively impact segment-matching performance. Tinchev \etal~\cite{2019-RAL-LTST} utilize convolution to obtain an embedding space suitable for urban and natural scenarios. They subsequently estimate match quality through probabilistic geometric validation.
\subsubsection{Classification-based Methods}
Several works compute the category of segments by performing classification in the descriptor space, categorized into SegMap family and other methods.

\textbf{SegMap Family}. As shown in Figure \ref{fig:segments}(b), the pioneering work SegMap~\cite{2018-RSS-SegMap} incrementally clusters point clouds to create a global segment map. It employs segment-wise KNN retrieval with a data-driven descriptor extractor comprising three convolutional and two fully connected layers, then assigns a classification score using a fully connected network. It enables high compression rates in environment reconstruction and facilitates large-scale 3D LiDAR SLAM. Subsequently, researchers successfully integrate SegMap ~\cite{2018-RSS-SegMap} into LiDAR SLAM~\cite{2020-ICRA-LOL} and segment-based mapping framework~\cite{2021-IROS-Semsegmap}.

\textbf{Other Methods}. 
Wietrzykowski \etal~\cite{2021-IROS-OTDP} propose a DNN that learns visual context from synthetic LiDAR intensity images. They claim that using the latest LiDAR and ambient images can yield additional performance improvements. OneShot~\cite{2020-ICRA-OneShot} employs a range image-based method for segment extraction and a custom-tailored neural network to extract LiDAR-Vision descriptors.
\subsubsection{Similarity-based Methods}
Another approach constructs segment-based descriptors and assessed their similarities, thus combining the advantages of segments and global descriptors. As shown in Figure \ref{fig:segments}(c), Locus ~\cite{2021-ICRA-Locus} encodes topological and temporal information of segments to create a global descriptor using second-order pooling and nonlinear transformation. It avoids global map construction, achieving robustness to viewpoint changes and occlusions.

\subsection{Observations and Implications}\label{sec:segment-summary}
Traditional point cloud descriptors rely on low-level properties~\cite{2011-SSRR-Z-Projection,2015-IROS-FastHistogram,2018-ICRA-Delight,2020-ICRA-ISC,2010-ICRA-RPRF} to encode the point cloud, but local descriptors lack description ability, and global descriptors struggle with rotation and translation invariance. Fortunately, segments offer a good compromise between the two. Several observations are summarized as follows:

\textbf{(i)} Segments~\cite{2017-ICRA-Segmatch,2018-RSS-SegMap,2021-IROS-Semsegmap} offer a potential solution to reduce feature computation by avoiding processing the entire point cloud. Nevertheless, several approaches rely on point cloud aggregation or map construction, leading to inefficiencies when dealing with large-scale environments. \textbf{(ii)} Segment-based methods show promise in enhancing accuracy by incorporating geometric, color, and semantic information of segments. However, they require rich 3D geometry structures for segmentation, which may not always be available, thus limiting their applicability. \textbf{(iii)} Segment-based methods are well-known for their resilience to environmental changes, encompassing illumination, weather, and seasonal variations. However, they offer limited insights into the underlying 3D structures, resulting in subpar segmentation performance during long-term localization scenarios with numerous moving objects.
\section{LPR Techniques: Semantics}\label{sec:lpr-techniques-semantics}
Semantics refers to labels or categories that divide point clouds into various instances using learning-based segmentation technology, facilitating semantic-level place recognition. Thus, semantics-based place recognition falls under the category of learning-based methods. Based on the approach used for semantics association, they can be classified into two types: graph-based and graph-free.
\subsection{Learning-based Methods}\label{sec:semantics-learning-based-methods}
\subsubsection{Graph-based Methods}
Semantic graphs intuitively depict the location and topological information of objects. Graph similarity and graph matching are two typical graph operations. 
\begin{figure*}[tbp]
	\centering
	\subfigure[Graph-based Methods]{\includegraphics[width=4.8cm] {image/SGPR-v4.jpg}}
	\hspace{0.2cm}
	\subfigure[Graph-free Methods]{\includegraphics[width=4.8cm] {image/2021-IROS-SSC-2.jpg}}
	\caption{Two semantics-based methods. (a) and (b) are originally shown in ~\cite{2020-IROS-SGPR} and ~\cite{2021-IROS-SSC}, respectively.}\label{fig:semantics}
\end{figure*}

\textbf{Graph Similarity.} As depicted in Figure \ref{fig:semantics}(a), SGPR~\cite{2020-IROS-SGPR, 2021-ICRA-SA-LOAM} represents semantic categories and centroids of points as nodes, capturing node feature relations through edges. It develops a GNN-based graph network with node embedding, graph embedding, and graph-graph interaction to compute graph similarity. SGPR demonstrates robustness against occlusion and viewpoint changes, especially for reverse loops.

\textbf{Graph Matching}. GOSMatch~\cite{2020-IROS-GosMatch} introduces an object-based place recognition approach for urban environments, which employs graph descriptors for candidate search and vertex descriptors for one-to-one correspondence calculation. BoxGraph~\cite{2022-arXiv-BoxGraph} stores object shapes in vertices and simplifies place recognition to an optimal vertex assignment problem. It employs bounding boxes as appearance embeddings for vertex entities and extends them for pose estimation. 
\subsubsection{Graph-free Methods}
Other works avoid semantic graph construction and mainly fall into two categories: semantic descriptors and other methods.

\textbf{Semantics Descriptors}. As shown in Figure \ref{fig:semantics}(b), Semantic Scan Context (SSC)~\cite{2021-IROS-SSC} enhances SC~\cite{2018-IROS-ScanContext} by utilizing semantics instead of height. Object Scan Context (OSC)~\cite{2022-arXiv-OSC} improves SC~\cite{2018-IROS-ScanContext} by constructing the descriptor around uniformly distributed objects (e.g., street lights and trash cans). Seq-Ndt~\cite{2019-IROS-SeqNdt} extends the NDT-based histogram descriptor\cite{2009-JFR-NDT-AAL} by incorporating semantic information and utilizes the Kullback-Leibler (KL) divergence to measure similarity.
RINet~\cite{2022-RAL-RINet} develops a lightweight siamese network with convolution, down-sampling, and attention mechanisms to compute descriptor similarities. It prioritizes scene learning over point cloud orientation and is highly efficient, allowing for deployment on resource-constrained platforms.

\textbf{Other Methods}. Recent studies enhance semantic-based LPR through innovative technologies and theories, including multiple hypothesis trees~\cite{2019-RAL-MHSM}, siamese neural network~\cite{2022-AR-OverlapNet}, spherical convolution~\cite{2021-TITS-PSE-Match}, and neural tensor network~\cite{2022-PRL-SC-LPR}.
\subsection{Observations and Implications}\label{sec:semantics-summary}
Inspired by human perception, semantic-based methods utilize pre-defined knowledge databases to categorize objects and identify their topological relationships. However, these methods are still relatively new and immature because they required advanced semantic segmentation technology. Several key observations are summarized below:

\textbf{(i)} Graph-based methods~\cite{2020-IROS-SGPR, 2021-ICRA-SA-LOAM,2020-IROS-GosMatch} have streamlined point cloud comprehension but exhibited three limitations. Firstly, potential loss of specific features, like object size. Secondly, inability to differentiate between parts of the same category leading to information loss. Thirdly, computing metrics between two graphs remains NP-complete, hindering the precise distance calculation within a reasonable timeframe. \textbf{(ii)} Semantic labels outperform using only geometric features, offering more interpretable and intuitive results. They demonstrate greater resilience to occlusion and viewpoint changes, especially in reverse LCD. However, predefined semantic labels in test datasets are limited, failing to encompass various categories in real-life scenarios. \textbf{(iii)} In dynamic or cluttered environments, leveraging objects and their topological information can enhance recognition accuracy. These methods heavily rely on the outcomes of semantic segmentation, which may lead to poor performance in diverse scenarios. Despite these challenges, they hold promise in applications where traditional methods fall short.
\section{LPR Techniques: Trajectory}\label{sec:lpr-techniques-trajectory}
Trajectory information enables correlating current and recent historical scans for place recognition. Odometry (handcrafted) and sequence (learning) are two prominent methods for historical data.
\subsection{Odometry-based Methods}\label{sec:trajectory-odometry-methods}
SLAM systems with LCD modules often adopt handcrafted approaches, utilizing front-end poses or traditional registration techniques~\cite{tao2021novel,shi2024ransac} for place recognition to reduce system complexity. They can be further categorized into naive Euclidean distance, overlap ratio, and PCR-based test.
\subsubsection{Naive Euclidean Distance}\label{sec:naive-euclidean-distance} 
Comparing the Euclidean distance between real-time and historical poses enables rough loop closure detection. Some works use piecewise orientation functions~\cite{2019-ECMR-LCDC} and global factor graphs~\cite{2020-IROS-LIO-SAM} for pose similarity comparison, while others employ multi-sensor calibration and mapping~\cite{2020-GRSL-TE3C}.
\subsubsection{Overlap Ratio}\label{sec:overlap-ratio}
The overlap ratio can assess place similarity, with a higher value indicating closer proximity. S4-SLAM~\cite{2021-AR-S4-SLAM} stores historical poses using a kd tree and evaluates candidate loops based on overlap rate. It balances real-time performance and accuracy, demonstrating robustness even with limited feature points and high moving speeds. Mendes \etal~\cite{2016-SSRR-IBPG} utilize an overlap criterion to generate new keyframes and implement a graphical model layer over LiDAR odometry to reduce drifts through graph-level loop closing.

\subsubsection{PCR-based Test}\label{sec:pcr-based-test}
PCR techniques verify candidate loops using relative poses, such as standard ICP, point-to-line/plane ICP, and Generalized ICP (GICP). 

\textbf{Standard ICP}. IN2LAAMA~\cite{2020-TRO-IN2LAAMA} devises an offline probabilistic framework that identifies loop closures using poses and validates candidates with an ICP test~\cite{1992-TPAMI-ICP}, proficiently handling motion distortion without an explicit motion model.

\textbf{Point-to-Line/Plane ICP}. Lego-LOAM~\cite{2018-IROS-Lego-LOAM} compares historical scans with pose constraints and refines transformations with ICP~\cite{1992-TPAMI-ICP}. It is a pioneering work to incorporate LCD into LiDAR SLAM, making it well-suited for long-duration navigation tasks. LILO~\cite{2021-TAES-LILO} extends this idea to a LiDAR-IMU system. Other works enhance registration robustness using plane graphs~\cite{2020-RAL-LipMatch}, intuitive weighting~\cite{2020-IROS-LiTAMIN}, and KL divergence~\cite{2021-ICRA-LiTAMIN2}, respectively.

\textbf{GICP}. LAMP~\cite{2020-ICRA-LAMP} develops a multi-robot LiDAR SLAM system for challenging subterranean environments that utilizes GICP~\cite{2009-RSS-GICP} to register nearby scans and proposes an Incremental Consistent Measurement (ICM) set maximization to reject outlying loop closures. 
\subsection{Sequence-based Methods}\label{sec:trajectory-sequence-based-methods}
SeqSLAM~\cite{2012-ICRA-SeqSLAM} pioneers visual feature similarity comparisons over time to integrate sequence information and identify the best match within local sequences, showcasing exceptional performance in extreme environmental changes and providing insights for LiDAR-based solutions. The point cloud sequence matching incurs higher computational costs than image-based alternatives. Consequently, approaches such as scan matching and submap matching integrate neural networks with GPUs to enhance efficiency.
\begin{figure*}[!t]
	\centering
	\subfigure[Scan Matching]{\includegraphics[height=2.6cm] {image/seqlpd-v4.jpg}}
	\hspace{0.1cm}
	\subfigure[Submap Matching]{\includegraphics[height=2.6cm] {image/2020-IROS-SeqSphereVLAD-2.jpg}}
	\caption{Two sequence-based methods. (a) and (b) are originally shown in~\cite{2019-IROS-Seqlpd} and \cite{2020-IROS-SeqSphereVLAD}, respectively. }\label{fig:trajectory-sequence}
\end{figure*}
\subsubsection{Scan Matching}\label{sec:scan-matching}
As shown in Figure \ref{fig:trajectory-sequence}(a), SeqLPD ~\cite{2019-IROS-Seqlpd} employs LPD-Net ~\cite{2019-ICCV-LpdNet} for global descriptor extraction and selects super keyframes based on feature space distribution. It combines super keyframe-based coarse matching with the local sequence fine matching to improve detection accuracy and efficiency. The trained model can be directly applied in real-world scenarios without additional training, facilitating practical applications.
\subsubsection{Submap Matching}\label{sec:submap-matching}
As illustrated in Figure \ref{fig:trajectory-sequence}(b), SeqSphereVLAD~\cite{2020-IROS-SeqSphereVLAD} and Yin \etal~\cite{2022-TIE-SequenceMatch} utilize a spherical convolution module to extract orientation-equivariant local features across multiple layers of spherical perspectives. It effectively handles changing viewpoints and addresses large-scale SLAM challenges. FusionVLAD~\cite{2021-RAL-Fusionvlad} proposes a multi-view fusion network that encodes top-down and spherical-view features from the local map, enhancing feature combination through a parallel fusion module for end-to-end training. It is well-suited for large-scale mapping tasks with limited computation resources.
\subsection{Observations and Implications}\label{sec:trajectory-summary}
Traditional frame-to-frame comparison methods yield an intuitive similarity score but intend to degrade in closed, symmetric, and dynamic environments. The trajectory-based approaches incorporate both spatial and temporal information to address this limitation. Two observations are summarized as follows:

\textbf{(i)} LiDAR SLAM~\cite{2018-IROS-Lego-LOAM,2020-ICRA-LAMP} employs a straightforward LCD method based on pose proximity, followed by PCR for calculating relative transformations. Despite satisfactory results, two limitations remain. Cumulative errors affect the reliability of odometry poses in large-scale scenarios. Furthermore, the local optimality of PCR impeds the integration of loop constraints into global optimization. \textbf{(ii)} Sequence-based methods exhibit versatility since they effectively incorporate diverse place recognition techniques, such as local and global descriptors~\cite{2019-IROS-Seqlpd,2022-TIE-SequenceMatch}, semantics, and segments. While visual sequence-based methods have been well-studied, LiDAR-based approaches are still in the early stages. Furthermore, the expensive calculations required for matching and feature fusion restrict their practical applicability.
\section{LPR Techniques: Map}\label{sec:lpr-techniques-map}
Map-assisted methods provide global metric localization to achieve place recognition. They generally fall into two groups based on the map construction timing: offline and online maps. 
\subsection{Handcrafted Methods}\label{sec:map-handcrafted-methods}
\subsubsection{Offline Map-based Methods}
As the vehicle moves, the static offline map confines its motion within predefined boundaries. Handcrafted methods mainly involve five map types: feature, probability, point cloud, grid, and mesh.

\textbf{Feature Map}.
Dong \etal~\cite{2021-ECMR-ORIP} employ range image-based pole extraction to build a global map and utilize Monte Carlo Localization (MCL) to update particle weights based on pole matching. Shi \etal~\cite{2021-AGCS-LIFM} use RANSAC~\cite{1981-CACM-RANSAC} to extract walls from the offline map and online scans, applying point-to-point and point-to-line distance constraints to compute vehicle poses. 

\textbf{Probability Map}. Schmiedel \etal~\cite{2015-IROS-IRON} characterize surface patches in NDT maps using curvature and object shape. They match descriptors between online scans and the global map, apply RANSAC~\cite{1981-CACM-RANSAC} for outlier detection, and evaluate matches using a normalized inlier ratio.

\textbf{Point Cloud Map}. 
Xu \etal~\cite{2022-ISPRSJ-CSSC} introduce a cross-section shape context descriptor that describes spatial distribution using elevation and point density, improving recognition performance with two-stage similarity estimation and the nearest cluster distance ratio. Shi \etal ~\cite{2023-JAG-100FPS,2023-ISPRSJ-AFLP} create an offline map database with a kd tree to simulate vehicle orientations and develop a binary loss function to improve localization accuracy.

\textbf{Grid Map}. Aldibaja \etal~\cite{2020-ITSC-LCAM} convert LiDAR scans into image-like representations of road surfaces, incorporating elevation and irradiation data. They employ a shared ID-based XY correlation matrix to represent loop-closure events among map nodes, facilitating large-scale map processing and map-combiner event detection independent of the driving trajectory.

\textbf{Mesh Map}. Chen \etal~\cite{2021-ICRA-RILL} employ Poisson surface reconstruction to generate a mesh map, developing an observation model of an MCL framework. It showcases robust generalization across different LiDAR sensors, eliminating the need for additional training data in varying environments.
\subsubsection{Online Map-based Methods}
SLAM dynamically constructs and updates an online point cloud map of the surrounding environment as the vehicle navigates within unknown terrain. MULLS~\cite{2021-ICRA-MULLS} incorporates TEASER~\cite{2021-TRO-TEASER} for loop verification and employs map-to-map ICP~\cite{1992-TPAMI-ICP} to enhance inter-submap edges with accurate transformations. CT-ICP~\cite{2022-ICRA-CT-ICP} projects a local map onto an elevation image, estimates a 2D transformation using RANSAC ~\cite{1981-CACM-RANSAC}, and computes a 6-DoF pose through ICP~\cite{1992-TPAMI-ICP} to identify potential loop closures. Liu \etal~\cite{2018-IV-R6LS} introduce a real-time 6D SLAM for large-scale natural terrains, which combines rotation histogram matching with a branch and bound search-based ICP~\cite{1992-TPAMI-ICP} to achieve real-time LCD. 
\subsection{Learning-based Methods}\label{sec:map-learning-based-methods}
\subsubsection{Offline Map-based Methods}
Researchers explore four types of offline maps in learning-based map localization: intensity, point cloud, node, and OpenStreetMap (OSM).

\textbf{Intensity Map}. Barsan \etal~\cite{2018-PMLR-LTLU} embed LiDAR intensity maps and online scans into a joint space to determine the vehicle's position. This method achieves centimeter-level accuracy and showcases robustness in handling uncalibrated data.

\textbf{Point Cloud Map}. Some works~\cite{2020-TITS-LocNet,2020-ICRA-Localising-Faster} employ DNN for feature learning and achieve global localization through the MCL framework. They address the non-conjugate issue between the Gaussian model and MCL, enhancing long-term localization performance. L3-Net~\cite{2019-CVPR-L3Net} captures temporal motion dynamics using deep Recurrent Neural Networks (RNNs), achieving comparable localization accuracy to SOTA methods. Retriever~\cite{2022-ICRA-Retriever} aggregates compact features with a perceiver for place recognition, enhancing computation efficiency by avoiding computation-heavy decompression.

\textbf{Node Map}. S4-SLAM2~\cite{2022-JIRS-S4-SLAM2} constructs a node map comprising point cloud, feature vectors, and location information. It extracts geometric and statistical features to create multi-modal descriptors and classified loop closures with a random forest classifier.

\textbf{OpenStreetMap}. OSM offers comprehensive geographic details such as streets, railways, water systems, and buildings. Several approaches~\cite{2017-ICRA-GONO,2019-ECMR-O4SD} integrates semantic information extracted from OSM into a particle filter framework. Cho \etal~\cite{2022-RAL-OBLG} generate a descriptor by calculating the distances to buildings at regular angles. 
\subsubsection{Online Map-based Methods}
These methods are roughly divided into surfel-based and grid map-based methods.

\textbf{Surfel Map}. SuMa~\cite{2018-RSS-SuMa} and SuMa++~\cite{2019-IROS-Suma++} employ range images and surfel-based maps for data association, detecting candidate loops by combining radius search and frame-to-model ICP~\cite{1992-TPAMI-ICP}. They verify loops by tracking poses, which ensures robust detection even with low overlap.

\textbf{Grid Map}. Yin \etal~\cite{2018-IROS-SAUF} generate a BEV map from the local occupancy map, considering vehicle motion errors. Furthermore, they introduce an additional GAN ~\cite{2016-CVPR-GAN} with conditional entropy reduction to enhance unsupervised feature learning for long-term recognition applications.
\subsection{Observations and Implications}\label{sec:map-summary}
Maps~\cite{2019-CVPR-L3Net,2020-ICRA-Localising-Faster,2023-JAG-100FPS} have been widely used in robot localization and path planning as they offer precise and detailed representations of the environment. Remarkably, map-based methods excel in recognizing topologically similar localization, providing pose information, and recovering kidnapped robots effectively. Several observations are summarized as follows: 

\textbf{(i)} Map representations enhance global consistency and reduced localization errors. However, their large memory requirements result in time-consuming loading, communication, and processing. \textbf{(ii)} Map can overcome noise and partial occlusions, ensuring robust recognition even in challenging scenarios. However, the significant density difference poses difficulties in registering online scans to maps. \textbf{(iii)} A robust prior map facilitates long-term localization in a consistent environment. However, significant environmental changes can cause the existing map to be outdated, resulting in localization errors.
\section{LPR Techniques: Other Methods}\label{sec:other-methods}
InCloud~\cite{2022-IROS-Incloud} distills the angular relationship between global representations, preserving the complex structure of the embedding space between training steps. Granström \etal~\cite{2010-IROS-LCL3} encode the point cloud using geometry features and range histograms, detecting loops with a trained classifier. While achieving high precision and recall rates, it requires the ordered point cloud.
\section{Datasets and Metrics}\label{sec:benchmarking}
\begin{table*}[!t]
	\tiny
	\centering
	\caption{A summary of existing datasets for LPR.}
	\begin{tabular}{|c|m{2.3cm}<{\raggedleft}|m{0.4cm}<{\centering}|m{0.8cm}<{\centering}|c|m{0.8cm}<{\centering}|m{2.8cm}<{\centering}|c|c|c|c|}
		\hline
		\textbf{Year}&\textbf{\makecell[c]{Name}}&\textbf{Seq}&\textbf{Trajectory (KM)}&\textbf{Type}&\textbf{\makecell[c]{Sensor\\Modilty}}&\textbf{Model of LiDAR}&\textbf{Loop}&\textbf{GT}&\textbf{LT}&\textbf{Public}\\\hline
		2009&Hannover2~\cite{2009-ICRA-NDT-ALD}&1&1.24&Out&L&&S+O&\checkmark&&\checkmark\\\hline
		2010&Freiburg~\cite{2010-ICRA-RPRF}&1&0.723&Out&L&SICK LMS&S&\checkmark&&\checkmark\\\hline
		2011&Ford Campus~\cite{2011-IJRR-FordCampus}&&&Out&C+L&Velodyne HDL-64E&&\checkmark&&\checkmark\\\hline
		2012&KITTI Odometry~\cite{2012-CVPR-KITTI}&22&39.2&Out&C+L&Velodyne HDL-64E&S+O&\checkmark&&\checkmark\\\hline
		2016&NCLT~\cite{2016-IJRR-NCLT}&27&147.4&In+Out&C+L&Velodyne HDL-32+Hokuyo UTM-30LX+Hokuyo URG-04LX&S+O&\checkmark&\checkmark&\checkmark\\\hline
		2017&Oxford RobotCar~\cite{2017-IJRR-Oxford-RobotCar}&$\textgreater$ 130&$\textgreater$1000&Out&C+L&SICK LD-MRS+SICK LMS-151&S+O&\checkmark&\checkmark&\checkmark\\\hline
		2018&Complex Urban~\cite{2018-ICRA-ComplexUrban}&19&158.82&Out&L&Velodyne VLP-16+SICK LMS-511&S+O&\checkmark&&\checkmark\\\hline
		2018&In-House~\cite{2018-CVPR-Pointnetvlad}&3&&Out&L&Velodyne HDL-64E&&\checkmark&\checkmark&\checkmark\\\hline
		2019&Semantic KITTI ~\cite{2019-ICCV-SemanticKITTI}&22&39.2&Out&L&Velodyne HDL-64E&S+O&\checkmark&&\checkmark\\\hline
		2019&Apollo-SouthBay ~\cite{2019-CVPR-L3Net}&&$\textgreater$380&Out&L&Velodyne HDL-64E&&\checkmark&\checkmark&\checkmark\\\hline
	    2019&HKUST~\cite{2019-arXiv-Fast-Complete}&&&In+Out&&Livox-MID40&&&&\checkmark\\\hline
		2020&MulRan~\cite{2020-ICRA-MulRan}&12&41.2&Out&L+R&Ouster OS1-64&S+O&\checkmark&\checkmark&\checkmark\\\hline
		2020&USyd~\cite{2020-ITSM-USYD}&$\textgreater$50&&Out&C+L&Velodyne VLP-16&S+O&\checkmark&\checkmark&\checkmark\\\hline
		2020&Oxford Radar Robotcar~\cite{2020-ICRA-Oxford-Radar-RobotCar}&$\textgreater$32&$\textgreater$280&Out&C+L+R&Velodyne HDL-32E+SICK LMS-151+Navtech CTS350-X&S+O&\checkmark&\checkmark&\checkmark\\\hline	
		2021&DUT-AS~\cite{2021-TIE-SIV}&30&&Out&L&SICK LMS 511&S+O&\checkmark&\checkmark&\\\hline
		2021&CMU Dataset~\cite{2021-TITS-PSE-Match}&11&2.0&Out&C+L&Velodyne VLP-16&&\checkmark&&\\\hline
		2021&Pittsburgh Dataset~\cite{2021-TITS-PSE-Match}&12&12.0&Out&C+L&Velodyne VLP-16&&\checkmark&&\\\hline
		2022&HAOMO ~\cite{2022-RAL-OverlapTransformer}&5&&Out&C+L&HESAI PandarXT-32&S+O&\checkmark&\checkmark&\checkmark\\\hline
		2022&Campus~\cite{2022-TIE-SequenceMatch}&11&2&Out&L&Velodyne-VLP 16&&\checkmark&&\\\hline
		2022&City~\cite{2022-TIE-SequenceMatch}&13&11&Out&L&Velodyne-VLP 16&&\checkmark&&\\\hline
		2022&KITTI-360~\cite{2022-TPAMI-KITTI-360} &9&73.7&Out&C+L&Velodyne HDL-64E&S+O&\checkmark&&\checkmark\\\hline
		2022&CHDloop~\cite{2022-IEEESensors-FSC}&5&1.519&Out&L&RoboSense RS-Ruby 128&S+O&\checkmark&&\\\hline
		2022&LGSVL~\cite{2022-TIE-MMCS-Net}&&&Out&&Velodyne HDL-64E&&\checkmark&\checkmark&\\\hline
		2022&Real Vehicle~\cite{2022-TIE-MMCS-Net}&&&Out&C+L&Velodyne VLP-32C&S+O&\checkmark&&\\\hline
	\end{tabular}
	\begin{tablenotes}
		\item \textbf{Seq}, \textbf{GT} and \textbf{LT} represent sequence, ground-truth, and long-term, respectively. \textbf{In} and \textbf{Out} mean indoor and outdoor, respectively. \textbf{C}, \textbf{L}, and \textbf{R} denote camera, LiDAR, and radar, respectively. \textbf{S} and \textbf{O} represent the same and oppo-direction loop, respectively.
	\end{tablenotes}
	\label{tab:sum-datasets}
\end{table*} 
\subsection{Datasets}\label{sec:bm-datasets}
A large number of datasets have been collected to evaluate the performance of LPR methods. Table \ref{tab:sum-datasets} provided a summary of these datasets. Their characteristics are summarized as follows:

\textbf{Long-term Collection}. ~\cite{2016-IJRR-NCLT,2017-IJRR-Oxford-RobotCar,2020-ICRA-Oxford-Radar-RobotCar} repeatedly gather the same scenario along similar routes in different seasons or times.

\textbf{Multi-modal Data}. In addition to LiDAR sensors, radar is used in ~\cite{2020-ICRA-MulRan,2020-ICRA-Oxford-Radar-RobotCar} and cameras are mounted in ~\cite{2011-IJRR-FordCampus,2012-CVPR-KITTI,2016-IJRR-NCLT,2017-IJRR-Oxford-RobotCar,2020-ITSM-USYD,2020-ICRA-Oxford-Radar-RobotCar,2021-TITS-PSE-Match,2022-RAL-OverlapTransformer,2022-TPAMI-KITTI-360}. Semantic information is also available in ~\cite{2019-ICCV-SemanticKITTI}.

\textbf{LiDAR Sparsity}. These datasets cover various density LiDAR sensors, such as mechanical 16-line~\cite{2018-ICRA-ComplexUrban,2020-ITSM-USYD}, 32-line~\cite{2016-IJRR-NCLT,2020-ICRA-Oxford-Radar-RobotCar}, 64-line~\cite{2011-IJRR-FordCampus, 2012-CVPR-KITTI}, and 128-line LiDAR~\cite{2022-IEEESensors-FSC}, as well as solid-state LiDAR~\cite{2019-arXiv-Fast-Complete}.

\textbf{Viewpoint Change}. In addition to same-direction revisits, ~\cite{2012-CVPR-KITTI,2016-IJRR-NCLT,2017-IJRR-Oxford-RobotCar,2018-ICRA-ComplexUrban,2019-ICCV-SemanticKITTI,2020-ICRA-MulRan,2020-ITSM-USYD,2020-ICRA-Oxford-Radar-RobotCar,2021-TIE-SIV,2022-RAL-OverlapTransformer,2022-TPAMI-KITTI-360,2022-IEEESensors-FSC,2022-TIE-MMCS-Net} contain reverse loops.

\textbf{Scenario Diversity}. These datasets are generally divided into two categories: indoor and outdoor datasets. Outdoor datasets are the most widely used, mainly including campuses ~\cite{2016-IJRR-NCLT}, highways~\cite{2018-ICRA-ComplexUrban}, rural areas~\cite{2012-CVPR-KITTI}, cities~\cite{2020-ITSM-USYD}, and riversides~\cite{2020-ICRA-MulRan}.
\subsection{Evaluation Metrics}\label{sec:bm-evaluation-metrics}
Different evaluation metrics have been proposed to test LPR methods, summarized as follows:

\textbf{Revisit Criteria}. A distance threshold is defined before evaluation to determine whether the query and candidate belong to the same place.

\begin{figure}[!t]
	\centering
	\subfigure[PR curve]{\includegraphics[width=3.4cm] {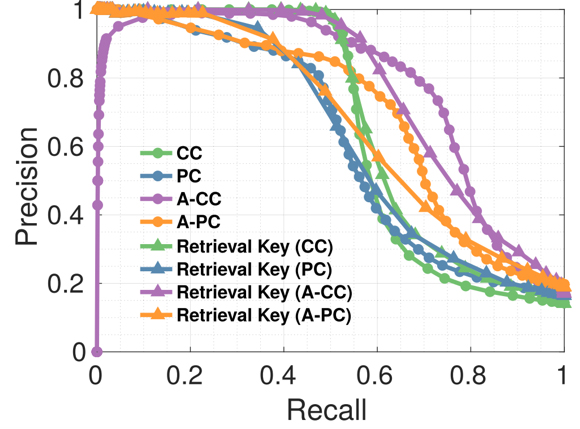}}
	\hspace{0.3cm}
	\subfigure[AUC]{\includegraphics[width=3.4cm] {image/auc-eps-converted-to.pdf}}
	\caption{An illustration of PR curve and AUC. (a) is originally shown in ~\cite{2021-TRO-ScanContext++}. The AUC in (b) corresponds to the area of the blue region.}\label{fig:metric}
\end{figure}

\textbf{Precision-recall (PR) Curves}~\cite{2012-TRO-DBoW2}. As depicted in Figure \ref{fig:metric}(a), this curve measures the relationship between Pecision (P) and Recall (R) under different threshold parameters. P measures the ratio of correct matches to the total of predicted positive instances, while R quantifies the proportion of real positive cases correctly identified as positive matches:
\begin{equation}
	\label{eq:precision}
	P = \frac{TP}{TP+FP}, \ R = \frac{TP}{TP+FN},
\end{equation}
where $TP$, $FP$, and $FN$ represent true positive, false positive, and false negative, respectively. 

\textbf{Area Under the PR Curve (AUC)}~\cite{2019-RAL-SCI}. As illustrated in Figure \ref{fig:metric}(b), it reflects the discrimination power of a LPR method and a larger AUC means more places are recognized with fewer errors. However, it does not retain any information regarding the features of the original PR Curve.

\textbf{Recall @Top-N}. It evaluates the accuracy of place recognition methods in identifying the correct places among the top-k retrieved matches. A higher value indicates better performance. TOP 1$\%$ ~\cite{2021-RAL-BVMatch} and TOP 1~\cite{2022-IROS-RING} are the two most frequently used metrics.

$F_\beta$ \textbf{Score}. It is the harmonic mean of precision and recall. A high value indicates the system struck a good balance between them as follows:  
\begin{equation}
	\label{eq:f-beta}
	F_\beta = (1+\beta^2)\times \frac{P \times R}{\beta^2 P + R},
\end{equation}
where $P$ and $R$ represent precision and recall, respectively. $\beta$ is a parameter that determines the weights of $P$ and $R$. $F_1$ score~\cite{2020-IROS-SGPR, 2022-AR-OverlapNet, 2021-IROS-SSC, 2021-PR-ATFF,2021-JAG-FastLCD, 2021-ICRA-Locus, 2022-arXiv-BoxGraph} is the most frequently used metric:
\begin{equation}
	\label{eq:f1}
	F_1 = 2 \times \frac{P \times R}{P + R},
\end{equation} 
where $F_1$ treats $P$ and $R$ as equally important. The maximum $F_1$ score ($F_1^m$) is then calculated as:
\begin{equation}
	\label{eq:f1-max}
	F_1^m = \underset{\tau}{max} \ 2 \times \frac{P_{\tau} \times R_{\tau}}{P_{\tau} + R_{\tau}},
\end{equation} 
where $\tau$ refers to a user-defined threshold for matching score or distance. For place recognition, we identify the best candidate for the query scan and compute their matching score or distance. We compare this score with $\tau$ to classify the result as $TP$, $FP$, or $FN$. We then calculate the precision, recall, and $F_1$ score for the entire sequence. Finally, we evaluate different $\tau$ values to determine the highest $F_1$ score as the final result $F_1^m$.

\textbf{Extended Precision (EP)}. It provides more comprehensive insights by simultaneously considering the lower and upper-performance bounds of an LPR method~\cite{2020-RAL-Metric-EP, 2021-IROS-SSC, 2021-ICRA-Locus}:  
\begin{equation}\label{eq:ep}
	EP = \frac{1}{2}\left(P_{R0} + R_{P100}\right),
\end{equation}
where $P_{R0}$ is the precision at minimum recall, and $R_{P100}$ is the max recall at 100$\%$ precision.

\textbf{Translation and Rotation Error}. In global localization-related tasks, the translation error $e_t$ quantifies the difference between the estimated and ground-truth translations, reflecting the accuracy of the robot's position. The rotation error $e_r$ measures the discrepancy between the estimated and actual rotations, indicating the accuracy of the robot's attitude.
\begin{equation}\label{eq:et-er}
	e_t = \boldsymbol{t}_{es} - \boldsymbol{t}_{gt},\ \ e_r = \arccos(trace(\mathbf{R}_{gt}^{\rm T}\mathbf{R}_{es})/2),
\end{equation}
where $\boldsymbol{t}_{es}$ and $\mathbf{R}_{es}$ denote the estimated translation and rotation, and $\boldsymbol{t}_{gt}$ and $\mathbf{R}_{gt}$ denote the ground-truth values. $trace$ refers to the matrix's trace.

\textbf{Localization Success Rate}. The success rate $rate_{success}$ is the proportion of successfully localized cases $N_{success}$ to the total cases $N_{total}$. A higher success rate indicates a more robust system capable of accurately localizing the robot from its initial state:
\begin{equation}\label{eq:success-rate}
	rate_{success} = \frac{N_{success}}{N_{total}},
\end{equation}
where a localization is successful only if the translation error $e_t$ is less than $\phi_t$ and the rotation error $e_r$ is less than $\phi_t$. $\phi_t$ and $\phi_t$ are thresholds for translation and rotation errors, respectively.

\textbf{Running Efficiency}. Runtime is crucial for online SLAM. Descriptor-based methods typically involve feature extraction and search, while map-based methods require map processing and matching. Data-driven methods, on the other hand, necessitate training and inference.
\section{Future Directions}\label{sec:future-directions}
\subsection{Multi-modality Information}
Multi-modality information offers the opportunity to leverage complementary features and enhance the robustness of localization. Several novel solutions are as follows:

\textbf{WiFi}. WiFi-based localization retrieves the Media Access Control (MAC) address of routers via the client device to calculate positions. This solution offers extensive coverage and achieves accurate indoor localization, overcoming GPS signal limitations. Moreover, it is easy to deploy and provides fast localization, holding significant potential for advancing indoor localization industries.

\textbf{Voice}. Voice-based localization uses microphone arrays to capture audio signals, employing delay estimation or spectral analysis to determine sound source locations. It offers advantages like low computational requirements, high concealment, and strong compatibility. Furthermore, it can seamlessly integrate with human-computer interaction systems, smart homes, and other voice-controlled applications, enhancing vehicle situational awareness.

\textbf{Radio Frequency Identification (RFID)}. RFID employs radio signals to identify and track objects without physical contact. Its anti-interference capability will ensure reliable autonomous driving in harsh environments. Additionally, the long lifespan can enhance the stability of long-term localization systems. Applying cryptographic encryption to tag data can strengthen system security.
\subsection{Innovative Solutions}
Breaking free from conventional solutions and incorporating interdisciplinary new technologies into autonomous driving holds the potential for unforeseen improvements: 

\textbf{Cloud Computing}. Cloud computing offers high-performance shared computing resources to users. Offloading computing tasks to cloud servers enhances robot localization efficiency. With access to powerful computing resources, robots can process diverse high-precision sensor data, thus improving their localization capabilities.

\textbf{Quantum Technology}. Quantum technology revolutionizes information calculation, encoding, and transmission. Quantum sensors can capture subtle changes, delivering ultra-high-precision measurements. Integrating such sensors into robotic navigation will greatly enhance localization and mapping in complex environments.

\textbf{Bio-inspired Localization}. Bio-inspired navigation and group behavior provide novel insights for robot localization. Inspired by turtles' navigation behavior, robots can enhance localization robustness by employing magnetic field sensors. Multi-robot systems can improve formation stability by emulating the cooperative behavior seen in bird flocks.
\subsection{Advanced Sensors}
Equipping robots with advanced sensors can enhance their navigation capabilities. We present several promising sensors as follows:

\textbf{Solid-State LiDAR}. Solid-state LiDAR employs a Micro-Electro-Mechanical System (MEMS), Optical Phased Array (OPA), or flash technology for signal transmission and reception. It boasts a compact size, high resolution, fast scanning speed, and extended measurement range. It enables precise identification of buildings, vehicles, and traffic signs, effectively ensuring autonomous driving safety, supporting smart transportation data, and monitoring traffic accidents.

\textbf{Event Camera}. Event cameras exclusively generate an asynchronous event stream when notable visual changes occur. They offer numerous benefits, including high time resolution, low latency, wide dynamic range, and low power consumption. Equipping vehicles with event cameras enhances obstacle avoidance in high-speed scenarios, enables navigation through scenes with abrupt light changes, and facilitates handling emergencies.

\textbf{Millimeter Wave Radar}. Millimeter-wave radar employs Frequency-Modulated Continuous Wave (FMCW) signals and mixers to measure speed, distance, and direction, providing cost-effectiveness, precise longitudinal ranging, accurate object detection, weather resistance, and high bandwidth. It will find wide applications in blind spot detection, object detection and tracking, parking assistance, and adaptive cruise control.
\subsection{Significant Applications}
As a new scientific and technological revolution unfolds, robot technology will spearhead advancements in several critical fields shaping the fate of humanity:

\textbf{Space Exploration}. In inhospitable environments like the moon, robots will assume the role of humans, undertaking tasks such as terrain mapping, mineral identification, house construction, and 3D printing. They will aid human understanding of deep space and other planets.

\textbf{Polar Research}. Robots advance polar research by enhancing data collection capabilities. They collect data on glaciers, weather, and temperature, facilitating continuous environmental monitoring. High-resolution mapping of polar topography helps identify landform changes, glacier movement, and geological processes.

\textbf{Underwater Robots}. Underwater robots are advanced submersibles tailored for extreme underwater operations. The fiber optic gyroscope and Doppler log will greatly enhance localization performance, benefiting port construction and naval defense. Sonar detection technology will further improve task efficiency like underwater rescue and pipeline maintenance.
\subsection{Approach Evaluations}
A fair and thorough evaluation is crucial for adapting robot products, algorithms, and scenarios. Here are three pivotal considerations for future algorithm evaluation.

\textbf{Scalability and Efficiency}. The growing affordability and accessibility of LiDAR sensors have spurred the demand for large-scale place recognition. This necessitates the development of scalable algorithms for handling large-scale point clouds and the design of efficient algorithms for the real-time processing of point cloud data on resource-constrained platforms.

\textbf{Long-term Place Recognition}. Long-term place recognition refers to the ability of a system to identify places over extended periods, despite appearance and weather variations. It is a crucial capability for autonomous navigation. Designing algorithms that can handle seasons, weather, appearances, and dynamic objects, will drive significant advancements in this field.

\textbf{Standardized Datasets}. A good dataset should possess sufficient size, high-quality data, reliable ground truth, annotations or labels, ethical considerations, normalized form, and clear instructions. Creating diverse data encompassing various sensor modalities, environmental conditions, and weather changes is also highly valuable.
%
%
%
\section{Conclusions}\label{sec:conclusions}
As high-level autonomous driving advances, robust navigation systems are essential for navigating complex environments. Place recognition enables vehicles to recognize previously visited locations despite changes in appearance, weather, and viewpoints, even determining their global location within prior maps. It is becoming increasingly important in autonomous driving. LiDAR, with its rich 3D data, long-range measurement, and stability in harsh conditions, has made LiDAR-based place recognition (LPR) a research hotspot. To address the gap in this field, we propose the first LPR survey. We first discuss the main problems solved by place recognition and analyze LPR's role in autonomous driving. Then, we offer a comprehensive problem statement, which divides LPR into implicit loop closure detection and explicit global localization to help readers understand the function of LPR in different requirements.

In recent years, many regions have implemented strict safety requirements for autonomous driving. LiDAR, as a reliable sensor, enhances system safety and reliability, making it an ideal choice. With technological advancements and increased production, LiDAR costs have decreased, broadening its application scope. Consequently, many LPR methods have emerged. We comprehensively classify these methods, describe their principles, and summarize their architectures, advantages, and disadvantages. These detailed analyses aim to help researchers understand each method's applicability and inspire advancements in place recognition technology for challenging environments such as docks, parks, and woodlands.

In the future, ongoing technological innovations will enhance LiDAR performance with higher resolution, faster scanning speeds, and longer detection distances. Large-scale production and usage of new materials will reduce costs, broadening LiDAR's applications. The fusion of LiDAR with other sensors (e.g., cameras, radars, GPS) will advance, providing more comprehensive and accurate localization and place recognition. To promote the further development of LPR, we summarize commonly used datasets, evaluation metrics, and promising future directions. Additionally, we maintain a project to collect SOTA LPR technologies, keeping the robotics community up to date.
\section{Acknowledgments} 
This work was supported by the National Natural Science Foundation of China (NSFC) under Grant 42030102 and 42271444, and the Science and Technology Major Project of Hubei Province under Grant 2021AAA010.
\bibliographystyle{unsrt}
\bibliography{REFs.bib}
\end{document}